\documentclass[10pt,journal,letterpaper,compsoc]{IEEEtran}
\usepackage{amsmath,amsfonts,amssymb,epsfig,verbatim,color}
\usepackage{graphicx}
\usepackage{multirow}
\usepackage{rotating}
\usepackage{booktabs}
\usepackage{pdfpages}
\usepackage{subfig}
\usepackage{algorithm}
\usepackage{algpseudocode}
\usepackage{url}
\usepackage{bm}
\usepackage{array}
\usepackage{xcolor,colortbl}
\usepackage{textcomp}
\newcommand{\mat}[1]{\mathtt #1}

\newcommand{\vct}[1]{\mathbf #1}
\newcommand{\T}{\ensuremath{^\top}}

\newcommand{\x}{{\bf x}}

\newcommand{\etal}{\textit{et al. }}

\DeclareMathAlphabet{\mathpzc}{OT1}{pzc}{m}{it}

\begin{document}
	
	\title{Large-scale Image Geo-Localization \\ Using Dominant Sets}
	
	\author{   Eyasu~Zemene*,~\IEEEmembership{Student Member,~IEEE,}
		Yonatan~Tariku*,~\IEEEmembership{Student Member,~IEEE,}
		Haroon~Idrees,~\IEEEmembership{member,~IEEE,
			Andrea~Prati,~\IEEEmembership{Senior member,~IEEE,}}
		Marcello~Pelillo,~\IEEEmembership{Fellow,~IEEE,}
		and~Mubarak~Shah,~\IEEEmembership{Fellow,~IEEE}% <-this % stops a space
		%The first and second authors have equal contribution.
		\IEEEcompsocitemizethanks{\IEEEcompsocthanksitem {* The first and second authors have equal contribution.}
			
			\IEEEcompsocthanksitem E. Zemene and M. Pelillo are with the department of Computer Science, Ca' Foscari University of Venice, Italy. E-mail: \{eyasu.zemene, pelillo\}@unive.it
			
			\IEEEcompsocthanksitem Y. Tariku Tesfaye is with the department of Design and Planning in Complex Environments of the University IUAV of Venice, Italy. E-mail: y.tesfaye@stud.iuav.it
			
			\IEEEcompsocthanksitem A. Prati is with the department of Department of Engineering and Architecture of the University of Parma, Italy. E-mail: andrea.prati@unipr.it
			
			\IEEEcompsocthanksitem M. Shah is with the Center for Research in Computer Vision (CRCV), University of Central Florida, USA. E-mail: \{haroon,shah\}@eecs.ucf.edu}}
	
	%IEEE TRANSACTIONS ON PATTERN ANALYSIS AND MACHINE INTELLIGENCE
	\markboth{}%
	{Shell \MakeLowercase{\textit{et al.}}: Bare Demo of IEEEtran.cls for Computer Society Journals}
	
	\IEEEcompsoctitleabstractindextext{
		\begin{abstract}
			
			This paper presents a new approach for the challenging problem of geo-localization  using image matching in a structured database of city-wide reference images with known GPS coordinates.  We cast the geo-localization as a clustering problem of local image features. Akin to existing approaches to the problem, our framework builds on low-level features which allow local matching between images. For each local feature in the query image, we find its approximate nearest neighbors in the reference set. Next, we cluster the features from reference images using Dominant Set clustering, which affords several advantages over existing approaches. First, it permits variable number of nodes in the cluster, which we use to dynamically select the number of nearest neighbors %(typically coming from multiple reference images) 
			for each query feature based on its discrimination value. Second, %as we also quantify in our experiments, 
			this approach is several orders of magnitude faster than existing approaches. Thus, we obtain  multiple clusters (different local maximizers) and obtain a robust final solution to the problem using multiple weak solutions through constrained Dominant Set clustering on global image features, where we enforce the constraint that the query image must be included in the cluster. This second level of clustering also bypasses heuristic approaches to voting and selecting the reference image that matches to the query. We evaluate the proposed framework on an existing dataset of 102k street view images as well as a new larger dataset of 300k images, and show that it outperforms the state-of-the-art by 20\% and 7\%, respectively, on the two datasets.
		\end{abstract}
		
		\begin{IEEEkeywords}
			Geo-localization, Dominant Set Clustering, Multiple Nearest Neighbor Feature Matching, Constrained Dominant Set
	\end{IEEEkeywords}}
	
	\maketitle
	
	\IEEEdisplaynotcompsoctitleabstractindextext \IEEEpeerreviewmaketitle
	
	%%%%%%%%% BODY TEXT
	
	\section{Introduction}
	\IEEEPARstart{I}{mage} geo-localization, the problem of determining the location of an image using just the visual information, is remarkably difficult. Nonetheless, images often contain useful visual and  contextual  informative cues which allow us to determine the location of an image with variable confidence. The foremost of these cues are landmarks, architectural details, building textures and colors, in addition to road markings and surrounding vegetation.
	
	Recently, the geo-localization through image-matching approach was proposed in \cite{zamir2010accurate, amirshahpami2014}. In \cite{zamir2010accurate}, the authors find the first nearest neighbor (NN) for each local feature in the query image, prune outliers and use a heuristic voting scheme for selecting the matched reference image. The follow-up work \cite{amirshahpami2014} relaxes the restriction of using only the first NN and proposed Generalized Minimum Clique Problem (GMCP) formulation for solving this problem. However, GMCP formulation can only handle a fixed number of nearest neighbors for each query feature. The authors used 5 NN, and found that increasing the number of NN drops the performance. Additionally, the GMCP formulation selects exactly one NN per query feature. This makes the optimization sensitive to outliers, since it is possible that none of the 5 NN is correct. Once the best NN is selected for each query feature, a very simple voting scheme is used to select the best match. Effectively, each query feature votes for a single reference image, from which the NN was selected for that particular query feature. This often results in identical number of votes for several images from the reference set. Then, both \cite{zamir2010accurate, amirshahpami2014} proceed with randomly selecting one reference image as the correct match to infer GPS location of the query image. Furthermore, the GMCP is a binary-variable NP-hard problem, and due to the high computational cost, only a single local minima solution is computed in \cite{amirshahpami2014}.
	
	In this paper, we propose an approach to image geo-localization by robustly finding a matching reference image to a given query image. This is done by finding correspondences between local features of the query and reference images. We first introduce automatic NN selection into our framework, by exploiting the discriminative power of each NN feature and employing different number of NN for each query feature. That is, if the distance between query and reference NNs is similar, then we use several NNs since they are ambiguous, and the optimization is afforded with more choices to select the correct match. On the other hand, if a query feature has very few low-distance reference NNs, then we use fewer NNs to save the computation cost. Thus, for some cases we use fewer NNs, while for others we use more requiring on the average approximately the same amount of computation power, but improving the performance, nonetheless. This also bypasses the manual tuning of the number of NNs to be considered, which can vary between datasets and is not straightforward. %Then, we prune query features that are not so informative to our geo-localization purposes.
	
	Our approach to image geo-localization is based on \textit{Dominant Set clustering} (DSC) - a well-known generalization of maximal clique problem to edge-weighted graphs- where the goal is to extract the most compact and coherent set. It's intriguing connections to evolutionary game theory allow us to use efficient game dynamics, such as replicator dynamics and infection-immunization dynamics (InImDyn). InImDyn has been shown to have a linear time/space complexity for solving standard quadratic programs (StQPs), programs which deal with finding the extrema of a quadratic polynomial over the standard simplex \cite{RotBomGEB2011,RotPelBomCVIU2011}. The proposed approach is on average 200 times faster and yields an improvement of 20\% in the accuracy of geo-localization compared to \cite{zamir2010accurate, amirshahpami2014}. This is made possible, in addition to the dynamics, through the flexibility inherent in DSC, which unlike the GMCP formulation avoids any hard constraints on memberships. This naturally handles outliers, since their membership score is lower compared to inliers present in the cluster. Furthermore, our solution uses a linear relaxation to the binary variables, which in the absence of hard constraints is solved through an iterative algorithm resulting in massive speed up.
	
	Since the dynamics and linear relaxation of binary variables allow our method to be extremely fast, we run it multiple times to obtain several local maxima as solutions. Next, we use a query-based variation of DSC to combine those solutions to obtain a final robust solution. The query-based DSC uses the soft-constraint that the query, or a group of queries, must always become part of the cluster, thus ensuring their membership in the solution. We use a fusion of several global features to compute the cost between query and reference images selected from the previous step. The members of the cluster from the reference set are used to find the geo-location of the query image. Note that, the GPS location of matching reference image is also used as a cost in addition to visual features to ensure both visual similarity and geographical proximity.
	
	GPS tagged reference image databases collected from user uploaded images on Flickr have been typically used for the geo-localization task. The query images in our experiments have been collected from Flickr, however, the reference images were collected from Google Street View. The data collected through Flickr and Google Street View differ in several important aspects: the images downloaded from Flickr are often redundant and repetitive, where images of a particular building, landmark or street are captured multiple times by different users. Typically, popular or tourist spots have relatively more images in testing and reference sets compared to less interesting parts of the urban environment. An important constraint during evaluation is that the distribution of testing images should be similar to that of reference images. On the contrary, Google Street View reference data used in this paper contains only a single sample of each location of the city. However, Street View does provide spherical $360^{\circ}$ panoramic views, , approximately 12 meters apart, of most streets and roads. Thus, the images are uniformly distributed over different locations, independent of their popularity. The comprehensiveness of the data ensures that a correct match exists; nonetheless, at the same time, the sparsity or uniform distribution of the data makes geo-localization difficult, since every location is captured in only few of the reference images. The difficulty is compounded by the distorted, low-quality nature of the images as well.

	The main contributions of this paper are summarized as follows: 
	\begin{itemize}
		%	\item We formulate image geo-localization in terms of a more generalized form of dominant sets framework. 
		%	\item We introduced a robust post processing step.
		%	\item We have collected new and more challenging high resolution reference dataset of 300K street view images. 
		\item We present a robust and computationally efficient approach for the problem of large-scale image geo-localization by locating images in a structured database of city-wide reference images with known GPS coordinates.   
		
		\item  We formulate geo-localization problem in terms of a more generalized form of dominant sets framework which incorporates weights from the nodes in addition to edges. %which affords several advantages. First, it permits variable number of nearest neighbors (NN) are selected dynamically based on their discriminative abilities. Second,
		
		\item We take a two-step approach to solve the problem. The first step uses local features to find putative set of reference images (and is therefore faster), whereas the second step uses global features and a constrained variation of dominant sets to refine results from the first step, thereby, significantly boosting the geo-localization performance.
		
		\item We have collected new and more challenging high resolution reference dataset (\textit{\textbf{WorldCities}} dataset) of 300K Google street view images.
		
	\end{itemize}
	
	The rest of the paper is structured as follows. We present literature relevant to our problem in Sec. \ref{secRelatedWork}, followed by technical details of the proposed approach in Sec. \ref{secFramework}, while constrained dominant set based post processing step is discussed in Sec. \ref{post-processing}. This is followed by dataset description  in section \ref{Dataset_discription}. Finally, we provide results of our extensive evaluation in Sec. \ref{secExperiments} and conclude in Sec. \ref{secConclusion}.

	\section{Related Work} \label{secRelatedWork}
	The computer vision literature on the problem of geo-localization can be divided into three categories depending on the scale of the datasets used: landmarks or buildings \cite{avrithis2010retrieving,chen2011city,quack2008world,zheng2009tour}, city-scale including streetview data \cite{jin2015predicting}, and worldwide \cite{hays2008im2gps,hays2015large,weyand2016planet}. Landmark recognition is typically formulated as an image retrieval problem \cite{avrithis2010retrieving,quack2008world, zheng2009tour,gammeter2009know,johns2011images}. For geo-localization of landmarks and buildings, Crandall \etal \cite{crandall2009mapping} perform structural analysis in the form of spatial distribution of millions of geo-tagged photos. This is used in conjunction with visual and meta data from images to geo-locate them. The datasets for this category contain many images near prominent landmarks or images. Therefore, in many works \cite{avrithis2010retrieving,quack2008world}, similar looking images belonging to same landmarks are often grouped before geo-localization is undertaken.
	
	For citywide geo-localization of query images, Zamir and Shah \cite{zamir2010accurate} performed matching using SIFT features, where each feature votes for a reference image. The vote map is then smoothed geo-spatially and the peak in the vote map is selected as the location of the query image. They also compute 'confidence of localization' using the Kurtosis measure as it quantifies the peakiness of vote map distribution. The extension of this work in \cite{amirshahpami2014} formulates the geo-localization as a clique-finding problem where the authors relax the constraint of using only one nearest neighbor per query feature. The best match for each query feature is then solved using Generalized Minimum Clique Graphs, so that a simultaneous solution is obtained for all query features in contrast to their previous work \cite{zamir2010accurate}. In similar vein, Schindler \etal \cite{SchindlerBS07} used a dataset of 30,000 images corresponding to 20 kilometers of street-side data captured through a vehicle using vocabulary tree. Sattler \etal \cite{SatHavSchPolCVPR2016} investigated ways to explicitly handle geometric bursts by analyzing the geometric relations between the different database images retrieved by a query. Arandjelovic´ \etal \cite{AraGroTorPajSicCVPR2016} developed a convolutional neural network architecture for place recognition that aggregates mid-level (conv5) convolutional features extracted from the entire image into a compact single vector representation amenable to efficient indexing. 
	Torii \etal \cite{TorSivOkuPajPAMI2015} exploited repetitive structure for visual place recognition, by robustly detecting repeated image structures and a simple modification of weights in the bag-of-visual-word model. Zeisl \etal \cite{BerTorMarICCV2015} proposed a voting-based pose estimation strategy that exhibits linear complexity in the number of matches and thus facilitates to consider much more matches.

	For geo-localization at the global scale, Hays and Efros \cite{hays2008im2gps} were the first to extract coarse geographical location of query images using Flickr collected across the world. Recently, Weyand \etal \cite{weyand2016planet} pose the problem of geo-locating images in terms of classification by subdividing the surface of the earth into thousands of multi-scale geographic cells, and train a deep network using millions of geo-tagged images. In the regions where the coverage of photos is dense, structure-from-motion reconstruction is used for matching query images \cite{agarwal2009building,li2012worldwide,sattler2011fast,sattler2012improving}. Since the difficulty of the problem increases as we move from landmarks to city-scale and finally to worldwide, the performance also drops.
	There are many interesting variations to the geo-localization problem as well. Sequential information such as chronological order of photos was used by \cite{kalogerakis2009image} to geo-locate photos. Similarly, there are methods to find trajectory of a moving camera by geo-locating video frames using Bayesian Smoothing \cite{vaca2012city} or geometric constraints \cite{hakeem2006estimating}. Chen and Grauman \cite{chen2011clues} present Hidden Markov Model approach to match sets of images with sets in the database for location estimation. Lin \etal \cite{lin2013cross} use aerial imagery in conjunction with ground images for geo-localization. Others \cite{lin2015learning,workman2015wide} approach the problem by matching ground images against a database of aerial images. Jacob \etal \cite{jacobs2007geolocating} geo-localize a webcam by correlating its video-stream with satellite weather maps over the same time period. Skyline2GPS \cite{ramalingam2010skyline2gps} uses street view data and segments the skyline in an image captured by an upward-facing camera by matching it against a 3D model of the city.
	
	Feature discriminativity has been explored by \cite{arandjelovic2014dislocation}, who use local density of descriptor space as a measure of descriptor distinctiveness, i.e. descriptors which are in a densely populated region of the descriptor space are deemed to be less distinctive. Similarly, Bergamo \etal \cite{bergamo2013leveraging} leverage Structure from Motion to learn discriminative codebooks for recognition of landmarks. In contrast, Cao and Snavely \cite{cao2013graph} build a graph over the image database, and learn local discriminative models over the graph which are used for ranking database images according to the query. Similarly, Gronat \etal \cite{gronat2013learning} train discriminative classifier for each landmark and calibrate them afterwards using statistical significance measures. Instead of exploiting discriminativity, some works use similarity of features to detect repetitive structures to find locations of images. For instance, Torii \etal \cite{torii2013visual} consider a similar idea and find repetitive patterns among features to place recognition. Similarly, Hao \etal \cite{hao20123d} incorporate geometry between low-level features, termed 'visual phrases', to improve the performance on landmark recognition.
	
	Our work is situated in the middle category, where given a database of images from a city or a group of cities, we aim to find the location where a test image was taken from. Unlike landmark recognition methods, the query image may or may not contain landmarks or prominent buildings. Similarly, in contrast to methods employing reference images from around the globe, the street view data exclusively contains man-made structures and rarely natural scenes like mountains, waterfalls or beaches.

	\begin{figure*}[t]%left,bottom,
		\centering
		\includegraphics[width=1\linewidth ,trim=0cm 1.8cm 0cm 2.3cm, clip]{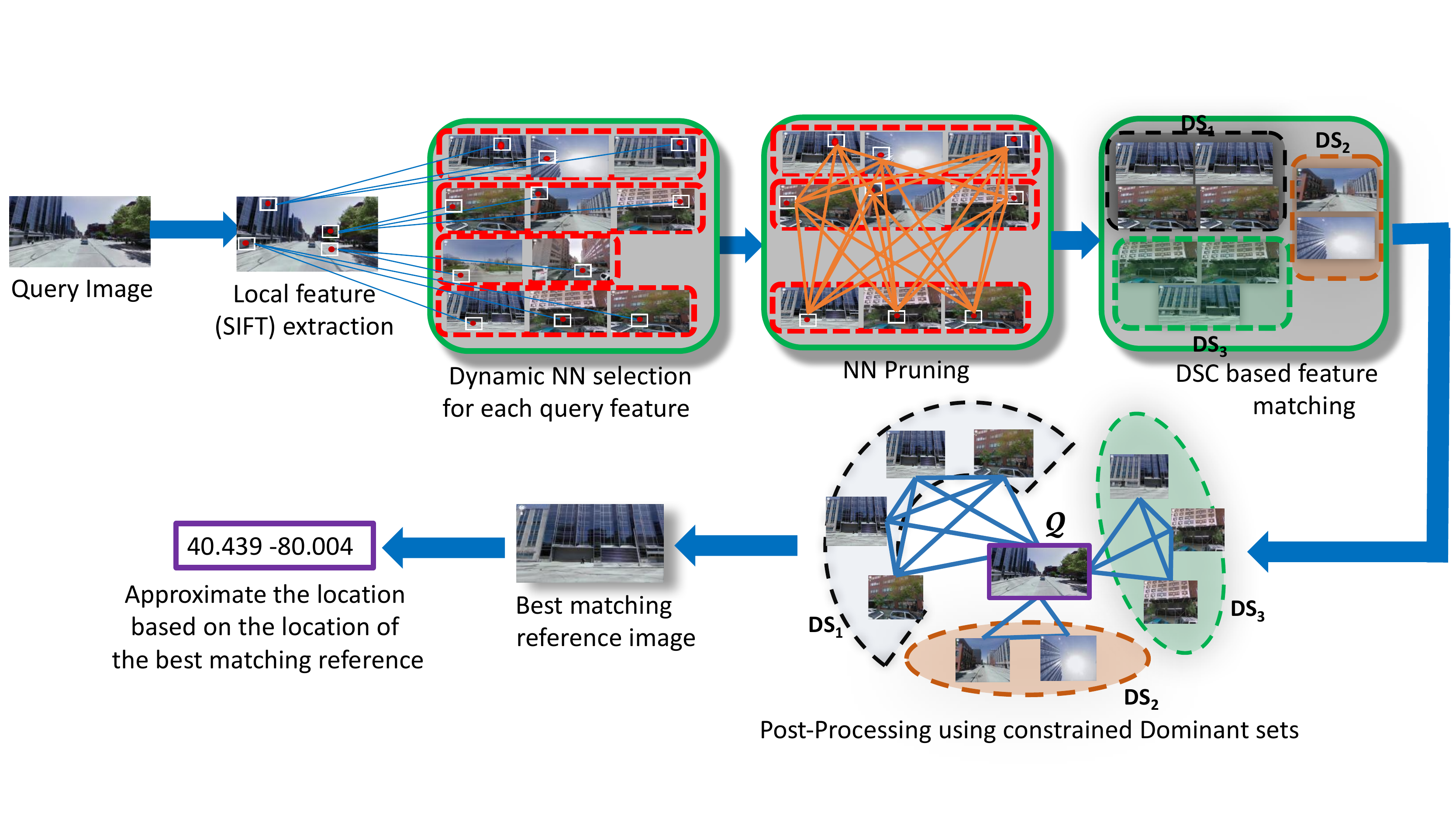}
		\caption{Overview of the proposed method.}
		\label{Overview}
	\end{figure*}
	\section{Image Matching Based Geo-Localization}\label{secFramework}
	Fig. \ref{Overview} depicts the overview of the proposed approach. Given a set of reference images, e.g., taken from Google Street View, we extract local features (hereinafter referred as \textit{reference features}) using SIFT from each reference image. We then organize them in a k-means tree \cite{DBLP:conf/visapp/MujaL09}.
	
	First, for each local feature extracted from the query image (hereinafter referred as \textit{query feature}), we dynamically collect nearest neighbors based on how distinctive the two nearest neighbors are relative to their corresponding query feature. Then, we remove query features, along with their corresponding reference features, if the ratio of the distance between the first and the last nearest neighbor is too large (Sec.~\ref{secAuto}). If so, it means that the query feature is not very informative for geo-localization, and is not worth keeping for further processing. In the next step, we formalize the problem of finding matching reference features to query features as a DSC (Dominant Set Clustering) problem, that is, selecting reference features which form a coherent and most compact set (Sec.~\ref{dsc-feature matching}). Finally, we employ constrained dominant-set-based post-processing step to choose the best matching reference image and use the location of the strongest match as an estimation of the location of the query image (Sec.~\ref{post-processing}).

	\subsection{Dynamic Nearest Neighbor Selection and Query Feature Pruning}\label{secAuto}
	For each of $N$ query features detected in the query image, we collect their corresponding nearest neighbors (\emph{NN}). Let $v_m^i$ be the $m^{th}$ nearest neighbor of $i^{th}$ query feature $q^i$, and $m \in \mathbb{N} : 1\leq m \leq \mid NN^i \mid$ and $i \in \mathbb{N} : 1 \leq i \leq N$, where $\mid\cdot\mid$ represents the set cardinality and $NN^i$ is the set of NNs of the $i^{th}$ query feature.
	% In \cite{amirshahpami2014} NN is fixed to 5, the downside of fixing the neighbors include with too small NN there will be high chance of missing important matching neighbor while too large NN will introduce outlier matching points. So,
	In this work, we propose a dynamic NNs selection technique based on how distinctive two consecutive elements are in a ranked list of neighbors for a given query feature, and employ different number of nearest neighbors for each query feature. %We use the following iterative formulation:
	%
	%\begin{equation}
	%% \begin{displaymath}
	%\begin{cases} \text{Add} \,\, \xi(v_{m+1}^i),&   \textrm{if} \; \frac{\parallel \xi(q^i)-\xi(v_m^i)\parallel}{\parallel \xi(q^i)-\xi (v_{m+1}^i) \parallel} >\tau,\\ \text{Stop},&\text{otherwise.}
	%\end{cases}
	%%\end{displaymath}
	%\label{ANN}
	%\end{equation}
	
	\begin{algorithm}[t]
		\caption{: Dynamic Nearest Neighbor Selection for $i^{th}$ query feature ($q^i$)}
		\label{alg:NNselection}
		
		\textbf{Input}: the $i^{th}$ query feature ($q^i$) and all its nearest neighbors extracted from K-means tree $\{ v_1^i,v_2^i.......v_{|NN^i|}^i\} $\\
		\textbf{Output}: Selected Nearest Neighbors for the $i^{th}$ query feature $(\mathbb{V}^i)$
		
		\rule{1\linewidth}{0.02cm}
		
		\begin{algorithmic}[1]
			\Procedure{Dynamic NN Selection}{$ $}
			\State Initialize $\mathbb{V}^i = \{v_1^i\}$ and m=1  
			\While {$m<|NN^i|-1$}
			\If {$\frac{\parallel \xi(q^i)-\xi(v_m^i)\parallel}{\parallel \xi(q^i)-\xi (v_{m+1}^i) \parallel} >\theta$ } 
			\State $\mathbb{V}^i=\mathbb{V}^i \cup v_{m+1}^i $ \Comment{If so, add $v_{m+1}^i$ to our solution }
			\State $m=m+1$ \Comment{Go to the next neighbor}
			\Else
			\State  Break \Comment{If not, stop adding and exit}
			\EndIf
			
			\EndWhile
			\EndProcedure
			%Return $\mathbb{V}$
		\end{algorithmic}
	\end{algorithm}
	As shown in Algorithm (\ref{alg:NNselection}), we add the $(m+1)^{th}$ NN of the $i^{th}$ query feature, $v_{m+1}^{i}$ , if the ratio of the two consecutive NN is greater than $\theta$, otherwise we stop. In other words, in the case of a query feature which is not very discriminative, i.e., most of its NNs are very similar to each other, the algorithm continues adding NNs until a distinctive one is found. In this way, less discriminative query features will use more NNs to account for their ambiguity, whereas more discriminative query features will be compared with fewer NNs.
	
	%\smallskip
	\noindent \textbf{Query Feature Pruning.} For the geo-localization task, most of the query features that are detected from moving objects (such as cars) or the ground plane, do not convey any useful information. If such features are coarsely identified and removed prior to performing the feature matching, that will reduce the clutter and computation cost for the remaining features. Towards this end, we use the following pruning constraint which takes into consideration distinctiveness of the \textit{first} and the \textit{last} NN. In particular, if  $\left\|\xi\left(q^i\right)-\xi\left(v_1^i\right)\right\| / \left\| \xi\left(q^i\right)-\xi\left(v_{\left|NN^i\right|}^i\right) \right\| > \beta$, where $\xi(\cdot)$ represents an operator which returns the local descriptor of the argument node, then $q^i$ is removed, otherwise it is retained. That is, if the \textit{first} NN is similar to the \textit{last} NN (less than $\beta$), then the corresponding query feature along with its NNs are pruned since it is expected to be uninformative. %$\parallel \cdot \parallel$ represents the distance between local features of the argument nodes. 
	
	%\smallskip
	We empirically set both thresholds, $\theta $ and $\beta$, in Algorithm (\ref{alg:NNselection})  and pruning step, respectively, to 0.7 and keep them fixed for all tests.

	\subsection{Multiple Feature Matching Using Dominant Sets}\label{dsc-feature matching}
	
	\subsubsection{The Dominant Set Framework}
	\label{Dominant set}
	
	The dominant set framework is a pairwise clustering approach \cite{PavPel07}, based on the notion of a dominant set, which can be seen as an edge-weighted generalization of a clique. The approach is a fast and efficient framework for pairwise clustering and has been used to solve multiple problems, such as data association in tracking \cite{YonEyaPelPraIET2016} as well as group detection \cite{VasMeqCriHunPelMurCVIU2015}.
	
	In an attempt to formally capture this notion, we present some notations and definitions.
	The data to be clustered is defined as a graph $G = (V,E,\zeta,\varpi)$, where $V, E,\zeta $ and $\varpi$ denote the set of nodes (of cardinality $n$), edges, node weights and edge weights, respectively. For a non-empty subset $S \subseteq V$, $l \in S$, and $k \notin S$, where $l$ and $k$ represent nodes in a graph $G$, we define $\phi_S(l,k)=\mat{B}(l,k)-\frac{1}{|S|} \sum_{p \in S} \mat{B}(l,p)$, where $\mat{B}$ is the corresponding $n \times n$ affinity matrix of graph $G$. This quantity measures the relative similarity between nodes $k$ and $l$, with respect to the average similarity between node $l$ and its neighbors in $S$. Note that $\phi_S(l,k)$ can be either positive or negative. Next, to each vertex $i \in S$ we assign a weight defined recursively as follows:
	\begin{equation}
	W_S(l)=
	\begin{cases}
	1,&\text{if\quad $|S|=1$},\\
	\sum_{k \in S \setminus \{l\}} \phi_{S \setminus \{l\}}(k,l)W_{S \setminus \{l\}}(k),&\text{otherwise}~.
	\end{cases}
	\end{equation}
	\noindent where $S \setminus \{l\}$ means set $S$ without the element $l$. Intuitively, $W_S(l)$ gives us a measure of the overall similarity between vertex $l$ and the vertices of $S\setminus \{l\}$,
	with respect to the overall similarity among the vertices in $S\setminus \{l\}$. Therefore, a positive $W_S(l)$ indicates that adding $l$ into its neighbors in $S$ will increase the internal coherence of the set, whereas in the presence of a negative value we expect the overall coherence to be decreased. The total weight of $S$ is computed as $W(S)=\sum_{l \in S}W_S(l)$.
	
	A non-empty subset of vertices $S \subseteq V$ such that $W(T) > 0$ for any non-empty $T \subseteq S$, is said to be a {\em dominant set} if:
	
	\begin{equation}
	\label{DSC_conditions}
	\begin{array}{ll}
	\text{$W_S(l) > 0,\forall l \in  S,$}  &    \\
	\text{$W_{S\cup \{l\}}(l) <0,\forall l \notin  S,$} &
	\end{array}
	\end{equation}
	
	These conditions correspond to the two main properties of a cluster: the first regards internal homogeneity, whereas the second regards external inhomogeneity.
	
\begin{figure}[t]
		\centering
		\includegraphics[width=1\columnwidth,trim=0cm 0.5cm 1cm 0cm, clip]{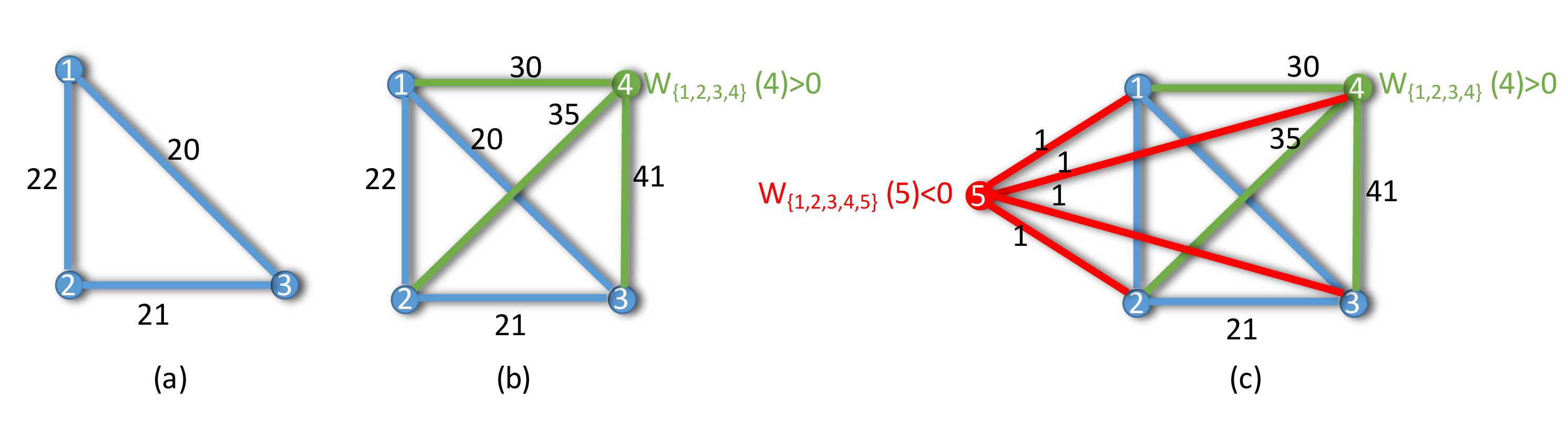}
		\caption{Dominant set example: (a) shows a compact set (dominant set), (b) node 4 is added which is highly similar to the set \{1,2,3\} forming a new compact set. (c) Node 5 is added to the set which has very low similarity with the rest of the nodes and this is reflected in the value $W_{\{1,2,3,4,5\}}(5)$.}
		\label{fig:exemplar}
	\end{figure}

	%\smallskip
	\textbf{Example:} Let us consider a graph with nodes $\{1,2,3\}$, which forms a coherent group (dominant set) with edge weights 20, 21 and 22 as shown in Fig. \ref{fig:exemplar}(a). Now, let us try to add a node $\{4\}$ to the graph which is highly similar to the set \{1,2,3\} with edge weights of  30, 35 and 41 (see Fig. \ref{fig:exemplar}(b)). Here, we can see that adding node \{4\} to the set increases the overall similarity of the new set \{1,2,3,4\}, that can be seen from the fact that the weight associated to the node \{4\} with respect to the set \{1,2,3,4\} is positive, $(W_{\{1,2,3,4\} }(4)>0)$. On the contrary, when adding node \{5\} which is less similar to the set \{1,2,3,4\} (edge weight of 1 - Fig. \ref{fig:exemplar}(c)) the overall similarity of the new set \{1,2,3,4,5\}  decreases, since we are adding to the set something less similar with respect to the internal similarity. This is reflected by the fact that the weight associated  to node \{5\} with respect to the set \{1,2,3,4,5\} is less than zero $(W_{\{1,2,3,4,5\}} (5)<0)$.
	
	%While adding node \{5\} which is less similar to the set \{1,2,3,4\} with edge weight of 1 see Figure \ref{fig:exemplar}(c). Unlike the previous case adding node \{5\} to the set will decrease the overall similarity of the new set \{1,2,3,4,5\}, since we are adding to the set something less similar with respect to the internal similarity. This is reflected by the fact that the weight associated to node \{5\} with respect to the set \{1,2,3,4,5\} is less than zero,$(W_{\{1,2,3,4,5\}} (5)<0).$
	
	From the definition of a dominant set in (\ref{DSC_conditions}) the set \{1,2,3,4\} (Fig. \ref{fig:exemplar} (b)) forms a dominant set, as it satisfies both criteria (internal coherence and external incoherence). % i.e. W{1,2,3,4} (1)>0, W{1,2,3,4} (2)>0, W{1,2,3,4} (3)>0, W{1,2,3,4} (4)>0 and a
	While the weight associated to the node out side of the set (dominant set) is less than zero, $W_{\{1,2,3,4,5\}} (5)<0$. %to node %\{5\} which is outside of the set (dominant set) is less than 0 (W{1,2,3,4,5} (5)<0).
	
	The main result presented in \cite{PavPel07} provides a one-to-one relation between dominant sets and strict local maximizers of the following standard quadratic optimization problem:
	
	%\begin{equation}
	%\label{quadratic_function}
	%\text{max} \quad X^T A X \quad \text{s.t.} \quad X \in \Delta
	%\end{equation}
	\begin{equation}
	\label{quadratic_function}
	\begin{array}{ll}
	\text{maximize }  &  f(\x) = \x\T \mat{B} \x, \\
	\text{subject to} &  \mathbf{x} \in \Delta,
	\end{array}
	\end{equation}
	\noindent where \[\Delta=\{ \x \in \mathbb{R}^n: \sum_i x_i = 1, \text{ and } x_i \geq 0 \text{ for all } i=1 \ldots n\},\]
	which is called \textit{standard simplex} of $\mathbb{R}^n$. Specifically, in \cite{PavPel07} it is shown that if $S$ is a dominant subset of vertices, then its weighted characteristic vector, which lies in $\vartriangle$,
	\begin{displaymath}
	x_l=
	\begin{cases} \frac{W_S(l)}{W(S)},&\text{if\quad $l \in S$},\\ 0,&\text{otherwise}
	\end{cases}
	\end{displaymath}
	is a strict local solution of (\ref{quadratic_function}). Conversely, under mild conditions, if $\x$ is a strict local solution of  (\ref{quadratic_function}), then its support $S = \sigma(\x) $ is a dominant set. Here, the support of a vector $\x\in\Delta $ is the set of indices corresponding to its positive components, that is $\sigma(\x)=\{l \in V : x_l >0\}$. By virtue of this result, a dominant set can be found by localizing a local solution of (\ref{quadratic_function}) and then picking up its support.
	
	\subsubsection{Similarity Function and Dynamics for Multiple Feature Matching} \label{similarity_dynm}
	
	In our framework, the set of nodes, $V$, represents all NNs for each query feature which survives the pruning step.
	The edge set is defined as $E=\{(v^i_m ,v^j_n) \mid  i\neq j \} $, which signifies that all the nodes in $G$ are connected as long as their corresponding query features are not the same. The edge weight, $\varpi : E \longrightarrow \mathbb{R}^+$ is defined as $\varpi (v^i_m, v^j_n) = exp(-\Vert \psi (v^i_m)  - \psi (v^j_n) \Vert^2/{2\gamma^2})$, where $\psi(\cdot)$ represents an operator which returns the global descriptor of the parent image of the argument node and $\gamma$ is empirically set to $2^7$ . The edge weight, $\varpi (v^i_m, v^j_n)$, represents a similarity between nodes $v^i_m $ and $v^j_n$ in terms of the global features of their parent images. The node score, $\zeta : V \longrightarrow \mathbb{R}^+$, is defined as $\zeta(v_m^i) = exp(-\Vert \xi(q^i)-\xi(v_m^i) \Vert^2/{2\gamma^2})$. The node score shows how similar the node $v^i_m$ is with its corresponding query feature in terms of its local features.
	
	Matching the query features to the reference features requires identifying the correct NNs from the graph $G$ which maximize the weight, that is, selecting a node (NN) which forms a coherent (highly compact) set in terms of both global and local feature similarities.

	Affinity matrix $\mat{A}$ represents the global similarity among reference images, which is built using GPS locations as a global feature and a node score $\vct{b}$ which shows how similar the reference image is with its corresponding query feature in terms of their local features. We formulate the following optimization problem, a more general form of the dominant set formulation:
	
	\begin{equation}
	\label{general_quadratic_function}
	\begin{array}{ll}
	\text{maximize }  &  f(\vct x) = \vct x\T\mat{A}\vct x + \vct b\T\vct x, \\
	\text{subject to} &  \mathbf{x} \in \Delta.
	\end{array}
	\end{equation}
	%where
	%\[
	%\Delta=\left\{ \vct x\in\mathbb R^\con n:\,\sum_{i=1}^\con n x_i=1 \text{ and } x_i\geq 0,\,i=1,\ldots,n\right\}\,.
	%\]
	
	The affinity $\mat{A}$ and the score $\vct{b}$ are computed as follows:
	\begin{align}
	\mat{A}(v^i_m, v^j_n)&=
	\begin{cases}
	\varpi (v^i_m, v^j_n)
	,&\text{for $i \neq j$,}\\
	0, &\text{otherwise,}
	\end{cases}\\
	%\]
	%and
	%\[
	\vct{b}(v_m^i)&= \zeta(v_m^i).
	\end{align}

	General quadratic optimization problems, like (\ref{general_quadratic_function}), are known to be NP-hard \cite{HorParTho2000}. However, in relaxed form, standard quadratic optimization problems can be solved using many algorithms which make full systematic use of data constellations. Off-the-shelf procedures find a local solution of (\ref{general_quadratic_function}), by following the paths of feasible points provided by game dynamics based on evolutionary game theory.
	
	Interestingly, the general quadratic optimization problem can be rewritten in the form of standard quadratic problem. A principled way to do that is to follow the result presented in \cite{Bom98}, which shows that maximizing the general quadratic problem over the simplex can be homogenized as follows. Maximizing $ \vct x\T\mat{A}\vct x + \vct b\T\vct{x}, \text{subject to } \vct{x} \in \Delta $ is equivalent to maximizing $\vct x\T\mat B\vct{x}, \text{subject to } \vct{x} \in \Delta$, where $\mat{B}=\mat{A}+\vct{e}\vct{b}\T+\vct{b}\vct{e}\T$ and $\vct{e}$ = $\sum\limits_{i=1}^{n}\vct{e}_i=[1,1,....1]$, where $\vct{e}_i$ denotes the $i^{th}$ standard basis vector in $\mathbb{R}^n$. This can be easily proved by noting that the problem is solved in the simplex.
	%\newpage
	
	\begin{itemize}
		\item[] $\vct x\T\mat{A}\vct x + 2*\vct b\T\vct x = \vct x\T\mat{A}\vct x + \vct b\T\vct x + \vct{x}\T\vct{b}$,
		\item[]\hspace{2.55cm}$= \vct{x}\T\mat{A}\vct{x} + \vct{x}\T\vct{e}\vct{b}\T\vct{x} + \vct{x}\T\vct{b}\vct{e}\T\vct{x}$,
		\item[]\hspace{2.55cm}$= \vct{x}\T(\mat{A} + \vct{e}\vct{b}\T + \vct{b}\vct{e}\T)\vct{x}$,
		\item[]\hspace{2.55cm}$= \vct{x}\T\mat{B}\vct{x}$.
	\end{itemize}
	
	As can be inferred from the formulation of our multiple NN feature matching problem, DSC can be essentially used for solving the optimization problem. Therefore, by solving DSC for the graph $G$, the optimal solution that has the most agreement in terms of local and global features will be found. Next, we introduce some useful dynamics from evolutionary game theory which allow us to efficiently and effectively match reference features to the query features.
	
	%\subsubsection{Dynamics for Multiple Feature Matching}
	
	The standard approach for finding the local maxima of problem (\ref{quadratic_function}), as used in \cite{PavPel07},
	is to use replicator dynamics - a well-known family of algorithms from evolutionary game theory inspired by Darwinian selection processes. Variety of StQP applications \cite{BomGO97,KarJanNeuCraGabJMRI2007,PavPel07,PelEO2009,SinNarIJCV2008} have proven the effectiveness of replicator dynamics. However, its computational complexity, which is $O(N^2)$ for problems involving $N$ variables, prevents it from being used in large-scale applications. Its exponential variant, though it reduces the number of iterations needed for the algorithm to find a solution, suffers from a per-step quadratic complexity.
	
	In this work, we adopt a new class of evolutionary game dynamics called infection-immunization dynamics (InImDyn), which have been shown to have a linear time/space complexity for solving standard quadratic programs. In \cite{RotPelBomCVIU2011}, it has been shown that InImDyn is orders of magnitude faster but as accurate as the replicator dynamics.
	
	The dynamics, inspired by infection and immunization processes summarized in Algorithm (\ref{alg:inimdyn}), finds the optimal solution by iteratively refining an initial distribution $\vct{x} \in \Delta$. The process allows for invasion of an infective distribution $\vct{y}\in \Delta$ that satisfies the inequality $ (\vct y - \vct x)\T\mat{B}\vct x > 0$, and combines linearly $\vct{x}$ and $\vct{y}$ (line 7 of Algorithm (\ref{alg:inimdyn})), thereby engendering a new population $\vct{z}$ which is immune to $\vct{y}$ and guarantees a maximum increase in the expected payoff.
	%Evolutionary game theory assumes a scenario where a non-rational pairs of players, which play based on a pre-assigned set of strategies, repeatedly drawn from a large population  plays a symmetric two-player game which drives the strategies with lower payoff to extinction.
	%Let $x_i(t)$ is the proportion of the population which plays strategy $i$ $\in J$(set of strategies) at time $t$. The state of the population at any given instant is then given by $\vct{x}_i(t)$ = [$x_1,x_2, ..., x_n$]$\T$ where $\T$ denotes transposition and $n$ refers the size of available pure strategies, $|J|$.
	%A distribution $\vct{y} \in \Delta $ is an infective strategy for another distribution $\vct{x}$, which is not local solution (equilibrium), if $ (\vct y - \vct x)\T\mat{A}\vct x > 0$ \cite{RotBomGEB2011}. % Hence, the set of infective distributions for $\vct x$ is given by\label{not:infective}
	%\[
	%\Upsilon(\vct x)=\{\vct y\in\Delta:\,(\vct y - \vct x)\T\mat{A}\vct x>0\}\,.
	%\]
	A selective function, $\mathcal{S}(\vct{x})$, returns an infective strategy for distribution $\vct{x}$ if it exists, or $\vct{x}$ otherwise (line 2 of Algorithm (\ref{alg:inimdyn})). %Intuitively, it returns a strategy between best performing pure strategy $r$ and worst one $r'$ which yields the largest expected payoff against $\vct{x}$. One can compute $r$ and $r'$ as follows: $r=\vct{e}^u$ where $u=argmax_{i\in J}(\mat{A}\vct{x})_j$ and $r'=\frac{x_v}{1-x_v}(\vct{x}-\vct{e}^v) + \vct{x}$ where $v=argmin_{i\in \sigma(\vct{x})}(\mat{A}\vct{x})_j$. The infective strategy $\vct{y}$ is then computed as $\vct{y}=\mathcal{S}(\vct{x}) \in argmax_{z\in\{r,r'\}}\vct z\T\mat{A}\vct z$.
	Selecting a strategy $\vct{y}$ which is infective for the current population $\vct{x}$, the extent of the infection, $\delta_{\vct y}(\vct x)$, is then computed in lines 3 to 6 of Algorithm (\ref{alg:inimdyn}).
	
	%\begin{equation}\label{eq:delta}
	%{\delta}_{\vct y}(\vct x)=
	%\begin{cases}
	% \min\left\{\frac{(\vct x - \vct y)\T\mat{A}\vct x}{{(\vct y - \vct x)\T\mat{A}(\vct y-\vct x)}},1\right\} \text{if}{(\vct %y-\vct x)\T\mat{A}(\vct y-\vct x)}<0\\
	%1 \text{, \hspace*{3cm}                           otherwise}\,.
	%\end{cases}
	%\end{equation}
	
	%Allowing for invasion, combining linearly $\vct{x}$ and $\vct{y}$, a new population $\vct{z}$ which is immune to $\vct{y}$ arises guarantee a maximum increase in the expected payoff.
	
	By reiterating this process of infection and immunization the dynamics drives the population to a state that cannot be infected by any other strategy. If this is the case then $\vct{x}$ is an equilibrium or fixed point under the dynamics. The refinement loop of Algorithm (\ref{alg:inimdyn}) controls the number of iterations allowing them to continue until $\vct{x}$ is with in the range of the tolerance $\tau$ and we emperically set $\tau$ to $10^{-7}$. The range $\epsilon(\vct{x})$ is computed as  $\epsilon(\vct{x})$ = $\sum\limits_{i\in J}\textrm{min}\left\{x_i,(\mat{B}\vct{x})_i-\vct{x}\T\mat{B}\vct{x}\right\}^2$.

	\begin{algorithm}[t]
		\caption{FindEquilibrium($\mat{B}$,$\vct x$,$\tau$)}\label{alg:inimdyn}
		\textbf{Input}:  $n\times n$ payoff matrix $\mat{B}$, initial distribution $\vct x\in\Delta$ and tolerance $\tau$.\\
		\textbf{Output}: Fixed point $\vct x$
		\begin{algorithmic}[1]
			\While {$\epsilon(\vct x)>\tau$} \\
			$\vct y\gets \mathcal S(\vct x)$ \\
			$\delta\gets 1$
			\If{${(\vct y-\vct x)\T\mat{B}(\vct y-\vct x)}<0$}
			\State $\delta\gets\min\left\{\frac{(\vct x - \vct y)\T\mat{B}\vct x}{{(\vct y - \vct x)\T\mat{B}(\vct y-\vct x)}},1\right\}$
			\EndIf
			\State $\vct x\gets \delta(\vct y-\vct x)+\vct x$
			\EndWhile\\
			\Return $\vct x$
		\end{algorithmic}
		\vspace{-.03in}
	\end{algorithm}
	
	As it can be inferred from the above formulation of dominant sets, finding the strict local solution of (\ref{quadratic_function}) coincides with finding the best matching reference features to the query features. It is important to note that, since our solution is a local solution and we do not know which local solution includes the best matching reference image, we determine several (typically three) locally optimal solutions. Unlike most of the previous approaches, which perform a simple voting scheme to select the best matching reference image, we introduce a post processing step utilizing a variant of dominant set called \textit{constrained dominant set}, which is discussed briefly in the next section.
	
	%As can be inferred from the formulation of our multiple NN feature matching problem, DSC can be essentially used for solving the same optimization problem. Therefore, by solving DSC for the graph $G$; the optimal solution which has the most agreement in terms of global and local features will be found.
	
	\section{Post processing using constrained dominant sets}\label{post-processing}
	
	Up to now, we devised a method to collect matching reference features corresponding to our query features. The next task is to select one reference image, based on feature matching between query and reference features, which best matches the query image. To do so, most of the previous methods follow a simple voting scheme, that is, the matched reference image with the highest vote is considered as the best match. %which ever reference image which belong to the reference features which have the most matches from will be considered as the best match.
	
	This approach has  two important shortcomings.  First, if there are  equal votes for two or more reference images (which happens quite often), it will randomly select one, which makes it prone to outliers. Second, a simple voting scheme does not consider the similarity between the query image and the candidate reference images at the global level, but simply accounts for local matching of features. Therefore, we deal with this issue by proposing a post processing step, which considers the comparison of the global features of the images and employs \emph{constrained dominant set}, a framework that generalizes the dominant sets formulation and its variant \cite{PavPel07,PavPel03}.
	%As shown in \cite{EyaPel16}, where the framework is applied to interactive image segmentation, by fixing a  regularization parameter one can determine the structure and the scale of the underlying problem and be able to extract groups of dominant set clusters which are constrained to contain user-specified query.
	
	\subsection{Constrained Dominant Sets Framework}\label{constraint_dom}
	
	In our post processing step, the user-selected query and the matched reference images are related using their {\em global} features and a unique local solution is then searched which contains the union of all the dominant sets containing the query. As customary, the resulting solution will be globally consistent both with the reference images and the query image and  due to the notion of \textit{centrality}, each element in the resulting solution will have a membership score, which depicts how similar a given reference image is with the rest of the images in the cluster. So, by virtue of this property of constrained dominant sets, we will select the image with the highest membership score as the final best matching reference image and approximate the location of the query with the GPS location of the best matched reference image.
	
	In this section, we review the basic definitions and properties of constrained dominant sets, as introduced in \cite{EyaPel161}.
	Given a user specified query, $\mathcal{Q} \subseteq \hat V$, we define the graph $\hat G=(\hat V,\hat E,\hat w)$, where the edges are defined as $\hat{E}=\{ (i,j) | i\neq j,  \{i,j\} \in \mathcal{DS}_n \vee  (i \in \mathcal{Q} \vee  j \in \mathcal{Q}) \}$, i.e., all the nodes are connected as long as they do not belong to different local maximizers, $\mathcal{DS}_n$, which represents the $n^{th}$ extracted dominant set. The set of nodes $\hat{V}$ represents all matched reference images (local maximizers) and query image, $\mathcal{Q}$. The edge weight $\hat{w}: \hat{E} \rightarrow\mathbb{R}^+$ is defined as:
	
	%\[
	$$
	\hat{w}(i,j)=
	\begin{cases}
	\rho(i,j),&\text{for $i \neq j$, $i \in \mathcal{Q} \vee  j\in \mathcal{Q}$},\\
	\mat{B}_n(i,j),&\text{for $i \neq j$, $\{i,j\} \in \mathcal{DS}_n$},\\
	%1,&\text{if $\mathcal{R}(i)$ = $\mathcal{R}(j)$ (Black edges)}.\\
	0, &\text{otherwise}
	\end{cases}$$
	%\],
	\noindent where $\rho(i,j)$ is an operator which returns the global similarity of two given images $i$ and $j$, that is, $\rho (i, j) = exp(-\Vert \psi (i)  - \psi (j) \Vert^2/{2\gamma^2})$, $\mat{B}_n$ represents a sub-matrix of $\mat{B}$, which contains only members of $\mathcal{DS}_n$, normalized by its maximum value and finally $\mat{B}_n(i,j)$ returns the normalized affinity between the $i^{th}$ and $j^{th}$ members of $\mathcal{DS}_n$. The graph $\hat{G}$ can be represented by an $n\times n$ affinity matrix $\hat{B} = (\hat{w}(i,j))$, where $n$ is the number of nodes in the graph.
	
	Given a parameter $ \alpha > 0$, let us define the following parameterized variant of program (\ref{quadratic_function}):
	\begin{equation}
	\label{eqn:parQP}
	\begin{array}{ll}
	\text{maximize }  &  f_{\mathcal{Q}}^\alpha(\x) = \x\T (\hat {\mat{B}} - \alpha \hat I_{\mathcal{Q}}) \x, \\
	\text{subject to} &  \mathbf{x} \in \Delta,
	\end{array}
	\end{equation}
	\noindent where $\hat I_{\mathcal{Q}}$ is the $n \times n$ diagonal matrix whose diagonal elements are set to 1 in correspondence to the vertices contained in $\hat V \setminus \mathcal{Q}$ (a set $\hat V$ without the element $\mathcal{Q}$) and to zero otherwise.

	%Basically, the function $f_S^\alpha$ is obtained from $f$ by inserting in the affinity matrix $A$, the value of the parameter $\alpha$ in the main diagonal positions corresponding to the elements of $V \setminus S$.

	Let $\mathcal{Q} \subseteq \hat V$, with $\mathcal{Q} \neq \emptyset$ and let $\alpha > \lambda_{\max}(\hat {\mat{B}}_{\hat V \setminus \mathcal{Q}}) $, where $\lambda_{\max}(\hat {\mat{B}}_{\hat V \setminus \mathcal{Q}})$ is the largest eigenvalue of the principal submatrix of $\hat{\mat{B}}$ indexed by the elements of $\hat V \setminus \mathcal{Q}$.
	If $\x$ is a local maximizer of $f_{\mathcal{Q}}^\alpha$ in $\Delta$, then
	$\sigma(\x) \cap \mathcal{Q} \neq \emptyset$. A complete proof can be found in \cite{EyaPel161}.
	
	The above result provides us with a simple technique to determine dominant-set clusters containing user-specified query vertices. Indeed, if $\mathcal{Q}$ is a vertex selected by the user, by setting
	\begin{equation}
	\label{alphabound}
	\alpha > \lambda_{\max}(\hat{\mat{B}}_{\hat V \setminus \mathcal{Q}}),
	\end{equation}
	we are guaranteed that all local solutions of (\ref{eqn:parQP}) will have a support
	that necessarily contains elements of $\mathcal{Q}$.
	
	%\subsection{Post Processing}

\begin{figure}[t]
		\includegraphics[width=1\linewidth,trim=0cm 0cm 0cm 0cm, clip]{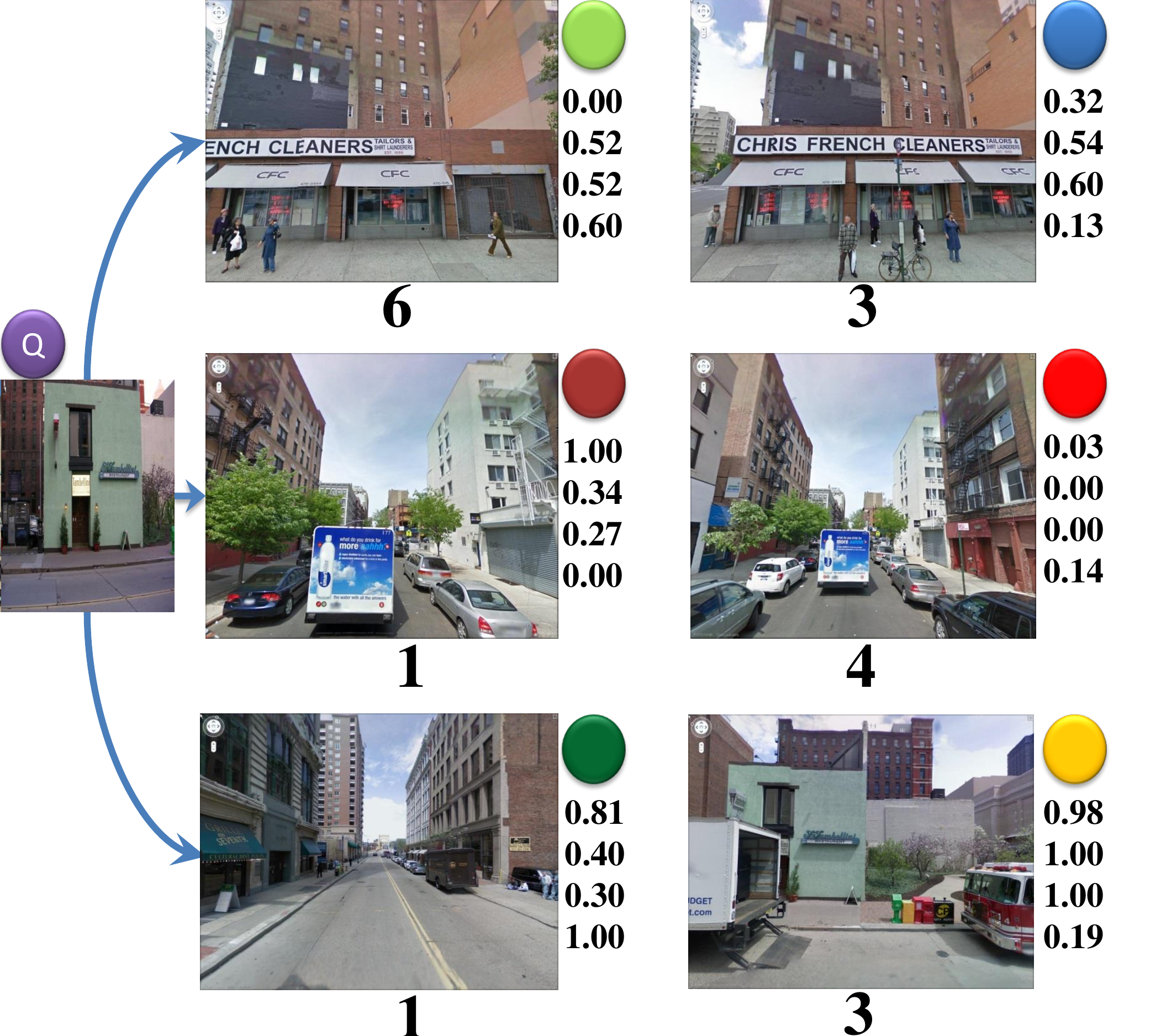}
		\caption{Exemplar output of the dominant set framework: {\bf Left:} query, {\bf Right:} each row shows corresponding reference images from the first, second and third local solutions (dominant sets), respectively, from top to bottom. The number under each image shows the frequency of the matched reference image, while those on the right side of each image show the min-max normalized scores of HSV, CNN6, CNN7 and GIST global features, respectively. The filled colors circles on the upper right corner of the images are used as reference IDs of the images. }
		\label{fig: Postprocessing1}
	\end{figure}

\begin{figure}[t]
		\includegraphics[width=1\linewidth,trim=0cm 0cm 0cm 0cm, clip]{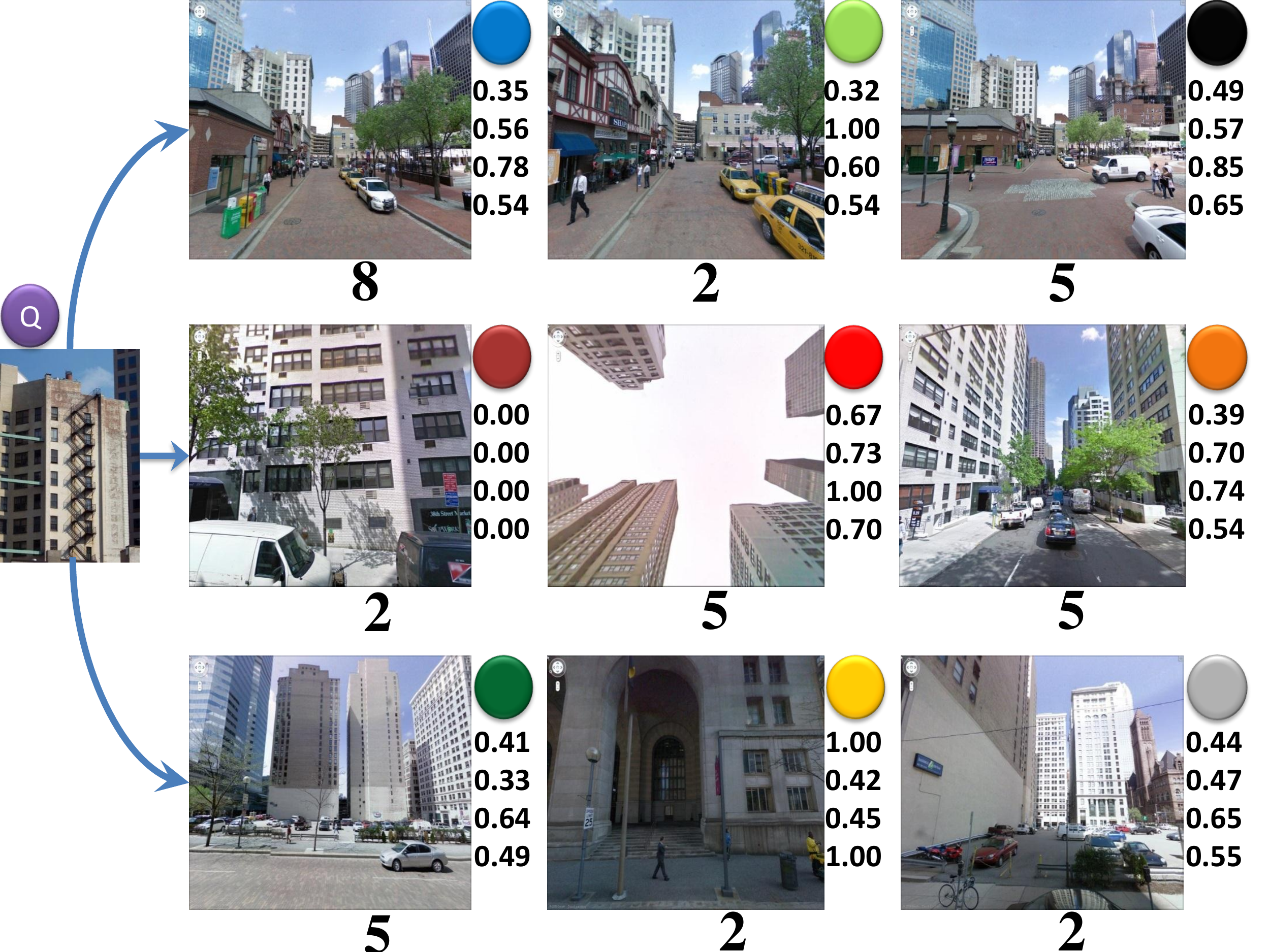}
		\caption{Exemplar output of the dominant set framework: {\bf Left:} query, {\bf Right:} each row shows corresponding reference images of the first, second and third local solutions (dominant sets), respectively, from top to bottom. The number under each image shows the mode of the matched reference image, while those on the right of each image show the min-max normalized scores of HSV, CNN6, CNN7 and GIST global features, respectively.}
		\label{fig: Postprocessing2}
	\end{figure}

	%=================================================================================
	The performance of our post processing may vary dramatically among queries and we do not know in advance which global feature plays a major role. Figs. \ref{fig: Postprocessing1} and \ref{fig: Postprocessing2} show illustrative cases of different global features. In the case of Fig. \ref{fig: Postprocessing1}, HSV color histograms and GIST provide a wrong match (dark red node for HSV and dark green node for GIST), while both CNN-based global features (CNN6 and CNN7) matched it to the right reference image (yellow node). The second example, Fig. \ref{fig: Postprocessing2}, shows us that only CNN6 feature localized it to the right location while the others fail. Recently, fusing different retrieval methods has been shown to enhance the overall retrieval performance \cite{LiaSheLuFeiZiQi15,ShaMinTimKaiDim15}.

	Motivated by \cite{LiaSheLuFeiZiQi15} we dynamically assign a weight, based on the effectiveness of a feature, to all global features based on the area under normalized score between the query and the matched reference images. The area under the curve is inversely proportional to the effectiveness of a feature. More specifically, let us suppose to have $\mathcal{G}$ global features and the distance between the query and the $j^{th}$ matched reference image ( $\mathcal{N}_j$), based on the $i^{th}$ global feature ($\mathcal{G}_i$), is computed as: $f_i^j=\psi_i(\mathcal{Q})-\psi_i(\mathcal{N}_j)$, where $\psi_i(\cdot)$ represents an operator which returns the $i^{th}$ global descriptor of the argument node. % the distance score between the query feature and the $i^{th}$ global feature $\mathcal{G}_i$ is $f_i=\psi(\mathcal{Q})-\psi(\mathcal{N}_i)$ \todo{The second one, shouldn't be $\mathcal{G}_i$ instead of $\mathcal{N}_i$?}.
	Let the area under the normalized score of $f_i$ be $\mathcal{A}_i$. The weight assigned for feature $\mathcal{G}_i$ is then computed as $w_i=\frac{1}{\mathcal{A}_i} / \sum\limits_{j=1}^{\mathcal{\left|G\right|}}\frac{1}{\mathcal{A}_j}$.

	Figs. \ref{fig: Postprocessing1} and \ref{fig: Postprocessing2} show illustrative cases of some of the advantages of having the post processing step. Both cases show the disadvantage of localization following heuristic approaches, as in \cite{zamir2010accurate,amirshahpami2014}, to voting and selecting the reference image that matches to the query. In each case, the matched reference image with the highest number of votes (shown under each image) is  the first node of the first extracted dominant set, but represents a wrong match. Both cases (Figs. \ref{fig: Postprocessing1} and \ref{fig: Postprocessing2}) also demonstrate that the KNN-based matching may lead to a wrong localization. For example, by choosing HSV histogram as a global feature, the KNN approach chooses as best match the dark red node in Fig. \ref{fig: Postprocessing1} and the yellow node in Fig. \ref{fig: Postprocessing2} (both with min-max value to 1.00). Moreover, it is also evident that choosing the best match using the first extracted local solution (i.e., the light green node in Fig. \ref{fig: Postprocessing1} and blue node in Fig. \ref{fig: Postprocessing2}), as done in \cite{amirshahpami2014}, may lead to a wrong localization,  since one cannot know in advance which local solution plays a major role. In fact, in the case of Fig. \ref{fig: Postprocessing1}, the third extracted dominant set contains the right matched reference image (yellow node), whereas in the case of Fig. \ref{fig: Postprocessing2} the best match is contained in the first local solution (the light green node).
	
	The similarity between query $\mathcal{Q}$ and the corresponding matched reference images is computed using their global features such as HSV histogram, GIST \cite{OlivTorIJCV2001} and CNN (CNN6 and CNN7 are Convolutional Neural Network features extracted from ReLU6 and FC7 layers of pre-trained network, respectively \cite{SimZisCoRR2014}). For the different advantages and disadvantages of the global features, we refer interested readers to \cite{amirshahpami2014}.
	
	%Since these global features and their advantages or disadvantages are not the main focus of this paper, we will not discuss them further. Interested readers can refer to \todo{Cite one or more references. Maybe previous Zamir-Shah's works?}. (CNN6 and CNN7 are Convolutional Neural Network features extracted from Relu6 and Fc7 layers of pretrained network, respectively

\begin{figure}[t]
		\includegraphics[width=1\linewidth,trim=0cm 0cm 0cm 0cm, clip]{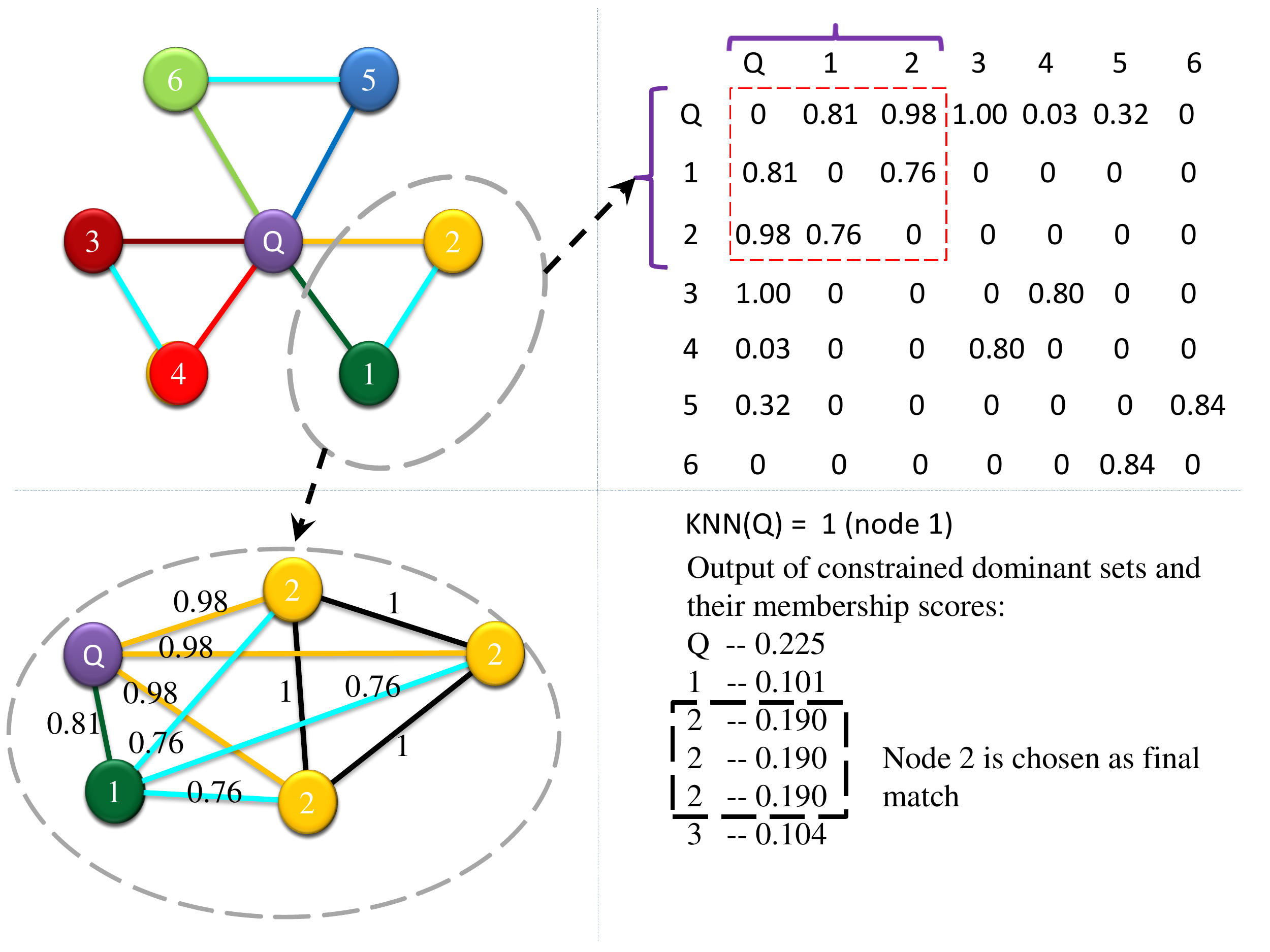}
		\caption{Exemplar graph for post processing. {\bf Top left:} reduced graph for Fig. \ref{fig: Postprocessing1} which contains unique matched reference images. {\bf Bottom left:} Part of the full graph which contains the gray circled nodes of the reduced graph and the query. {\bf Top right:} corresponding affinity of the reduced graph. {\bf Bottom right:} The outputs of nearest neighbor approach, consider only the node's pairwise similarity, (KNN($\mathcal{Q}$)=node 3 which is the dark red node) and constrained dominant sets approach ($\mathcal{CDS(Q)}$ = node 2 which is the yellow node).}
		\label{fig: PostprocessingGraph}
	\end{figure}

	Fig. \ref{fig: Postprocessing1} shows the top three extracted dominant sets with their corresponding frequency of the matched reference images (at the bottom of each image). Let $\mathcal{F}_i$ be the number  (cardinality) of local features, which belongs to $i^{th}$ reference image from the extracted sets and the total number of matched reference images be $\mathcal{N}$. We build an affinity matrix $\hat{\mat{B}}$ of size $\mathcal{S} = \sum\limits_{i=1}^{\mathcal{N}}\mathcal{F}_i + 1$ (e.g., for the example in Fig. \ref{fig: Postprocessing1}, the size $\mathcal{S}$ is 19 ). Fig. \ref{fig: PostprocessingGraph} shows the reduced graph for the matched reference images shown in Fig. \ref{fig: Postprocessing1}. Fig. \ref{fig: PostprocessingGraph} upper left, shows the part of the graph for the post processing. It shows the relation that the query has with matched reference images. The bottom left part of the figure shows how one can get the full graph from the reduced graph.
	%\end{equation}
	For the example in Fig. \ref{fig: PostprocessingGraph},  $\hat V=\{\mathcal{Q},1,2,2,2,3,4,4 .. ..... 6\}$.
	
	% Choose the regularizer $\alpha$ as $\lambda_{\max}(\hat{\mat{B}}_{\hat V \setminus \mathcal{Q}}) + \epsilon$  . Where $\epsilon$ is  a small positive number and $\lambda_{\max}(\hat{\mat{B}}_{\hat V \setminus \mathcal{Q}})$ is the largest eigenvalue of the principal submatrix of $\hat{\mat{B}}$indexed by the elements of $\hat V \setminus \hat S$ which means the affinity $\hat{\mat{B}}$ without the first row and column. If $\x$ is a local maximizer of $f_{\hat S}^\alpha$ in $\Delta$, then $\sigma(\x) \cap \mathcal{Q} \neq \emptyset$. We are guaranteed that the solution always contains the query.

	%Here, the similarity between the query and the matched reference image will be computed between their global features (CNN, HSV, RGB). while the between references will be 1 if the two references belong to the same cluster and 0 otherwise.
	
	The advantages of using constrained dominant sets are numerous. First, it provides a unique local (and hence global) solution whose support coincides with the union of all dominant sets of $\hat G $, which contains the query. Such solution contains all the local solutions which have strong relation with the user-selected query. As it can be observed in Fig. \ref{fig: PostprocessingGraph} (bottom right), the Constrained Dominant Set which contains the query $\mathcal{Q}$, $\mathcal{CDS(Q)}$, is the union of all overlapping dominant sets (the query, one green, one dark red and three yellow nodes) containing the query as one of the members. If we assume to have no cyan link between the green and yellow nodes, as long as there is a strong relation between the green node and the query, $\mathcal{CDS(Q)}$ will not be affected. In addition, due to the noise, the strong affinity between the query and the green node may be reduced, while still keeping the strong relation with the cyan link which, as a result, will preserve the final result. Second, in addition to fusing all the local solutions leveraging the notion of centrality, one of the interesting properties of dominant set framework is that it assigns to each image a score corresponding to how similar it is to the rest of the images in the solution. Therefore, not only it helps selecting the best local solution, but also choosing the best final match from the chosen local solution. Third, an interesting property of constrained dominant sets approach is that it not only considers the pairwise similarity of the query and the reference images, but also the similarity among the reference images. This property helps the algorithm avoid assignment of wrong high pairwise affinities. As an example, with reference to Fig. \ref{fig: PostprocessingGraph}, if we consider the nodes pairwise affinities, the best solution will be the dark red node (score 1.00). However, using constrained dominant sets and considering the relation among the neighbors, the solution bounded by the red dotted rectangle can be found, and by choosing the node with the highest membership score, the final best match is the yellow node which is more similar to the query image than the reference image depicted by the dark red node (see Fig. \ref{fig: Postprocessing1}).

	%\section{Experimental Results}\label{secExperiments}
	
	\begin{figure*}[t]
		\centering %trim=2cm(left) 3cm(bottom) 1.75cm(right) 2.5cm(top),clip
		\includegraphics[width=1\linewidth ,trim=0cm 1cm 0cm 0cm,clip]{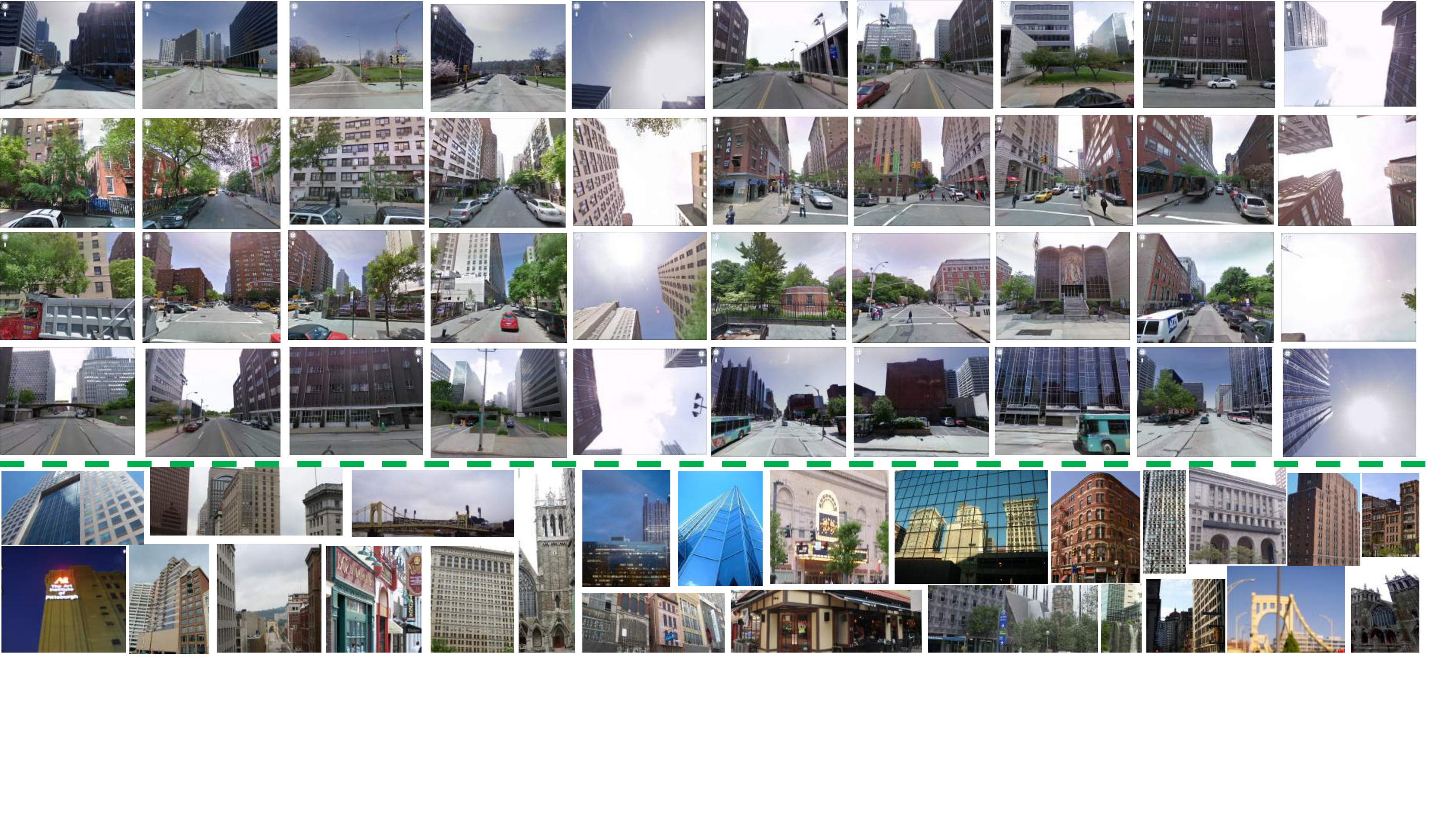}
		\vspace*{-2cm}
		\caption{The top four rows are sample street view images from eight different places of \textit{\textbf{WorldCities}} dataset. The bottom two rows are sample user uploaded images from the test set.}
		\label{sample_image}
	\end{figure*}
	
	\section{Experimental Results}\label{secExperiments}
	%In this section, we provide the details of our evaluation dataset and present our extensive experimental results for geo-localization and feature matching.
	
	%For all queries, we empirically set fixed value for all 3 parameters we used in our framework. Both $\tau $ and $\beta$ in section \ref{secAuto} are set to 0.7 and $\sigma$ in section \ref{constraint_dom} and subsection \ref{similarity_dynm} is set to $2^5$ and $2^7$, respectively. 
	
	\subsection{Dataset Description}\label{Dataset_discription}
	We evaluate the proposed algorithm  using publicly available reference data sets of over 102k Google street view images \cite{amirshahpami2014} and a new dataset, \textit{\textbf{WorldCities}}, of high resolution 300k Google street view images collected for this  work. The 360 degrees view of each place mark is broken down into one top and four side view images. The \textit{\textbf{WorldCities}} dataset is publicly available. \footnote{\url{http://www.cs.ucf.edu/~haroon/UCF-Google-Streetview-II-Data/UCF-Google-Streetview-II-Data.zip}}
	
	The 102k Google street view images dataset covers 204 Km of urban streets and the place marks  are approximately 12 m apart. It covers downtown and the neighboring areas of Orlando, FL; Pittsburgh, PA and partially Manhattan, NY.
	The \textit{\textbf{WorldCities}} dataset is a new high resolution reference dataset of 300k street view images that covers 14 different cities from different parts of the world: Europe (Amsterdam, Frankfurt, Rome, Milan and Paris), Australia (Sydney and Melbourne), USA (Vegas, Los Angeles, Phoenix, Houston, San Diego, Dallas, Chicago). Existence of similarity in buildings around the world, which can be in terms  of  their wall designs, edges, shapes, color etc, makes the dataset more challenging than the other. Fig. \ref{sample_image} (top four rows) shows sample reference images taken from different place marks.

	For the test set, we use 644 and 500 GPS-tagged user uploaded images downloaded from Picasa, Flickr and Panoramio for the 102k Google street view images and \textit{\textbf{WorldCities}} datasets, respectively. Fig. \ref{sample_image} (last two rows) shows sample test images. Throughout our experiment, we use the all the reference images from around the world to find the best match with the query image, not just with the ground truth city only.
	
	%In our experiments, each query image is matched against the entire reference data set and not the ground truth city only. Sample queries are shown in Figure \ref{sample_image}.

	%This section reports several experiments carried out to  analyze the various components of the proposed technique and provide quantitative results to show their robustness.
	
	%For all queries, we empirically set fixed value for all 3 parameters we used in our framework. Both $\tau $ and $\beta$ in section \ref{secAuto} are set to 0.7 and $\sigma$ in section \ref{constraint_dom} and subsection \ref{similarity_dynm} is set to $2^5$ and $2^7$, respectively.

	\subsection{Quantitative Comparison With Other Methods}
	\subsubsection{Performance on the 102k Google street view images Dataset}
\begin{figure}%bottom,right,top
		\centering
		\includegraphics[width=1\linewidth,trim=2.25cm 7.1cm 1.75cm 7.4cm, clip]{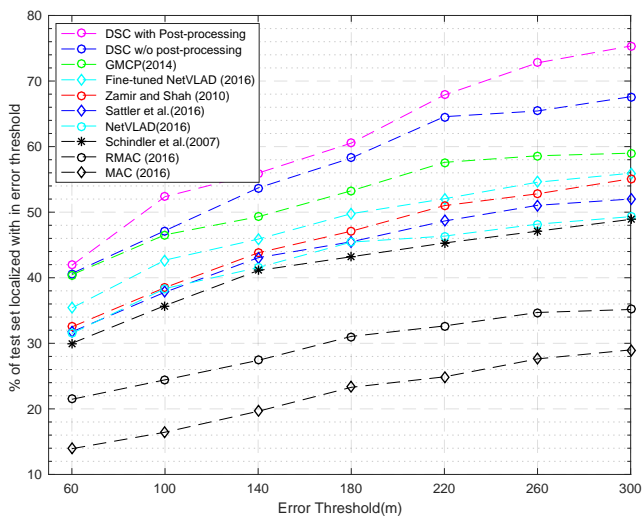}
		\caption{Comparison of our baseline (without post processing) and final method, on overall geo-localization results, with state-of-the-art approaches on the first dataset (102K Google street view images).}
		
		%and approaches proposed in \cite{amirshahpami2014, SchindlerBS07, TorSivOkuPajPAMI2015,ToliasICLR2016}}%Reported results of [zamir and shah] and [GMCP] and [schindler] are taken from [GMCP]
		\label{comparison_plot}
	\end{figure}

	The proposed approach has been then compared with the results obtained by state-of-the-art methods. In Fig. \ref{comparison_plot}, the horizontal axes shows the error threshold in meters and the vertical axes shows the percentage of the test set localized within a particular error threshold. Since the scope of this work is an accurate image localization at a city-scale level, test set images localized above 300 meter are considered a failure.
	
	The black (-*-) curve shows localization result of the approach proposed in \cite{SchindlerBS07} which uses vocabulary tree to localize images. The {\color {red} red (-o-)} curve depicts the results of \cite{zamir2010accurate} where they only consider the first NN for each query feature as best matches which makes the approach very sensitive to the query features they select. Moreover, their approach suffers from lacking global feature information. The {\color{green} green (-o-)} curve illustrates the localization results of \cite{amirshahpami2014} which uses generalized maximum clique problem (GMCP) to solve feature matching problem and follows voting scheme to select the best matching reference image. The black (-o- and $-\lozenge-$) curves show localization results of MAC and RMAC, (regional) maximum activation of convolutions (\cite{ToliasICLR2016,ToliasECCV2016}). These approaches build compact feature vectors that encode several image regions without the need to feed multiple inputs to the network. The {\color{cyan} cyan (-o-)} curve represents localization result of NetVLAD \cite{AraGroTorPajSicCVPR2016} which aggregates mid-level (conv5) convolutional features extracted from the entire image into a compact single vector representation amenable to effici,ent indexing. The {\color{cyan} cyan ($-\lozenge-$)} curve depicts localization result of NetVLAD but finetuned on our dataset. The {\color{blue} blue ($-\lozenge-$)} curve show localizaton result of approach proposed in \cite{SatHavSchPolCVPR2016} which exploits geometric relations between different database images retrieved by a query to handle geometric burstness. The {\color{blue} blue (-o-)} curve shows results from our baseline approach, that is,  we use voting scheme to select best match reference image and estimate the location of the query image. We are able to make a 10\% improvement w.r.t the other methods with only our baseline approach (without post processing). The {\color{magenta} magenta (-o-)} curve illustrates geo-localization results of our proposed approach using dominant set clustering based feature matching and constrained dominant set clustering based post processing. As it can be seen, our approach shows about 20\% improvement over the state-of-the-art techniques.

\begin{figure}
		\centering
		\includegraphics[width=1\linewidth,trim=4cm 8.5cm 3.75cm 9cm, clip]{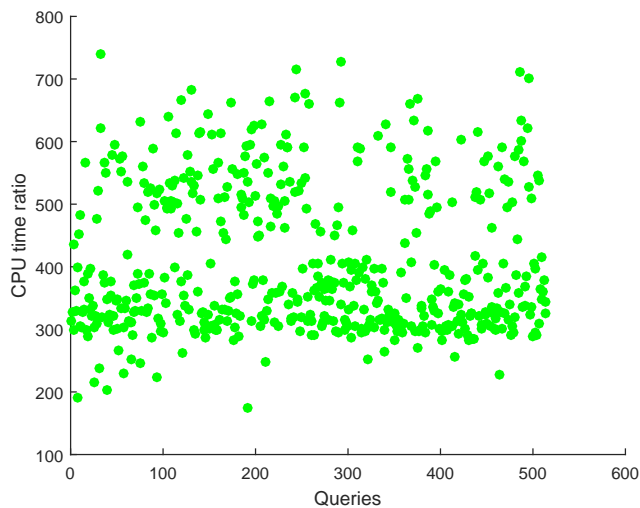}
		\caption{The ratio of CPU time taken between GMCP based geo-localization \cite{amirshahpami2014} and our approach, computed as CPU time for GMCP/CPU time for DSC.}
		\label{plot:cpuTimeRatio}
	\end{figure}

	\textbf{\textit{Computational Time.}} Fig. \ref{plot:cpuTimeRatio}, on the vertical axis, shows the ratio between GMCP (numerator) and our approach (denominator) in terms of CPU time taken for each query images to be localized. As it is evident from the plot, this ratio can range from 200 (our approach 200x faster than GMCP) to a maximum of even 750x faster.
	
	%\begin{figure}
	%	\centering
	%	\includegraphics[width=1\linewidth]{cpuTime2.png}
	%	\caption{CPU time taken for each query to be localized using our approach and GMCP-based geo-localization \cite{amirshahpami2014}.}
	%	\label{plot:time_plot}
	%\end{figure}

	\subsubsection{Performance on the WorldCities Dataset}
\begin{figure}
		\centering
		\includegraphics[width=1\linewidth,trim=2.00cm 7.1cm 1.75cm 7.4cm, clip]{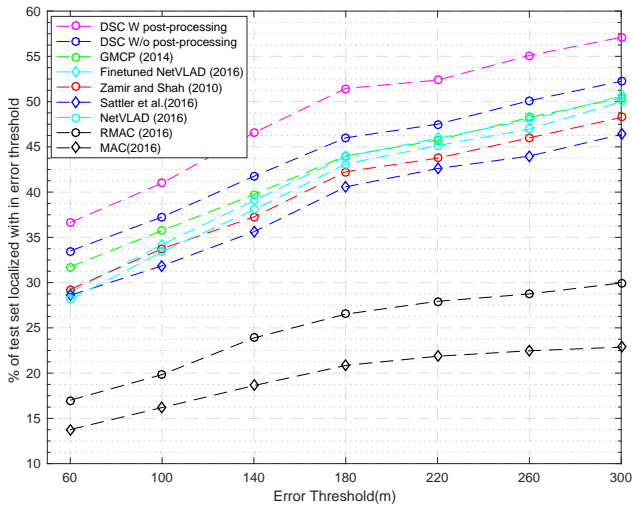}
		\caption{Comparison of overall geo-localization results using DSC with and without post processing  and  state-of-the-art approaches on the \textit{\textbf{WorldCities}} dataset. }%for GMCP \cite{amirshahpami2014} and \cite{zamir2010accurate}.}
		\label{comparisoin_plot_NewData}
	\end{figure}

	We have also compared the performance of different algorithms on the new dataset of 300k Google street view images created by us. Similarly to the previous tests, Fig.~\ref{comparisoin_plot_NewData} reports the percentage of the test set localized within a particular error threshold. Since the new dataset is relatively more challenging, the overall performance achieved by all the methods is lower compared to 102k image dataset.
	
	From bottom to top of the graph in Fig. \ref{comparisoin_plot_NewData} we present the results of \cite{ToliasICLR2016,ToliasECCV2016} black ($-\lozenge-$ and -o-), \cite{SatHavSchPolCVPR2016} {\color{blue} blue ($-\lozenge-$)}, \cite{zamir2010accurate} {\color {red} red (-o-)}, \cite{AraGroTorPajSicCVPR2016}  {\color{cyan} cyan (-o-)}, fine tuned \cite{AraGroTorPajSicCVPR2016} {\color{cyan} cyan ($-\lozenge-$)}, \cite{amirshahpami2014} {\color{green} green (-o-)}, our baseline approach without post processing {\color{blue} blue (-o-)} and our final approach with post processing  {\color{magenta} magenta (-o-)} . The improvements obtained with our method are lower than in the other dataset, but still noticeable (around 2\% for the baseline and 7\% for the final approach).
	
\begin{figure*}
		\centering
		\includegraphics[width=1\linewidth,trim=0cm 1.5cm 0cm 0.6cm, clip]{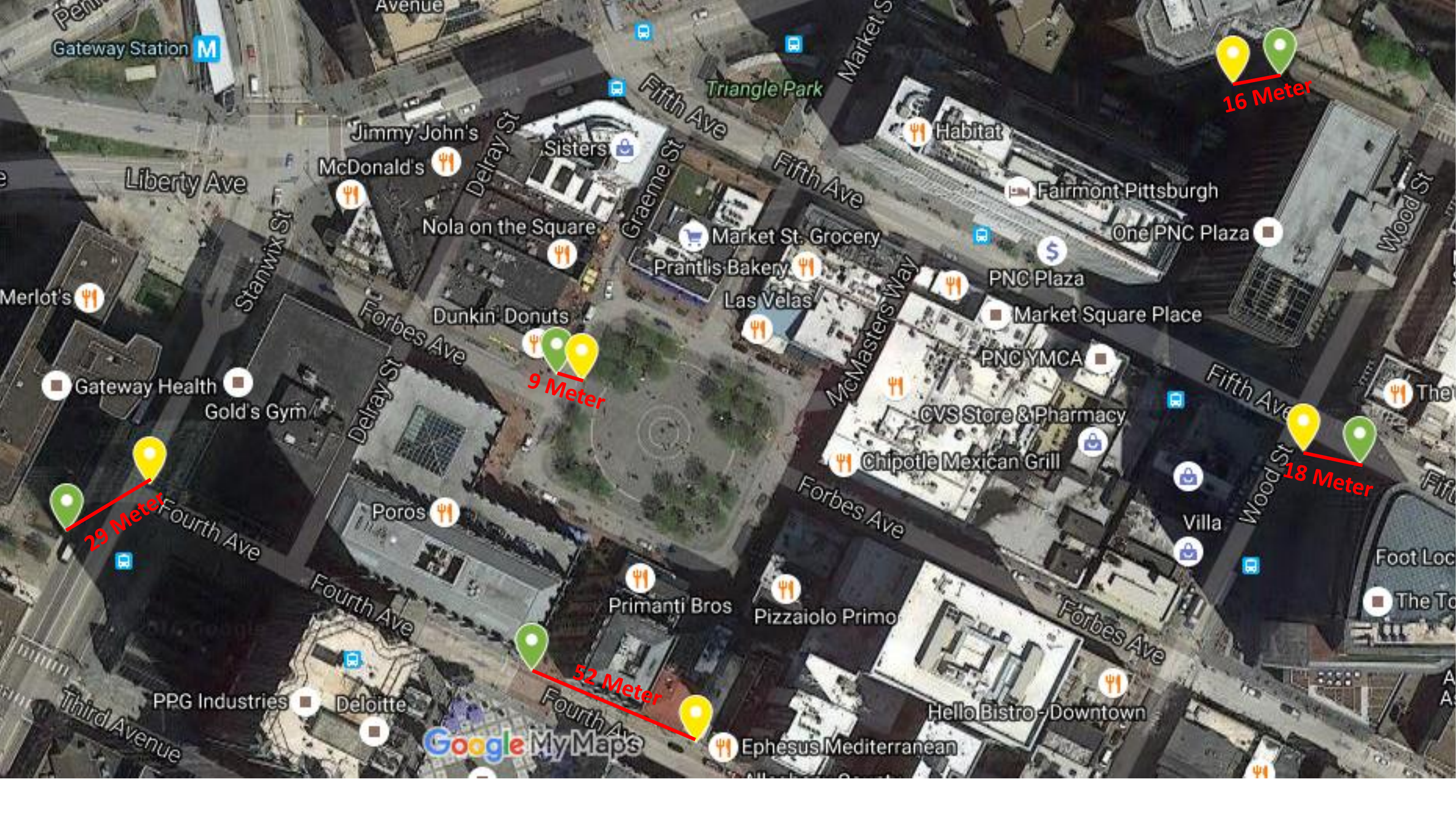}
		\vspace*{-0.5cm}
		\caption{Sample qualitative results taken from Pittsburgh area. The green ones are the ground truth while yellow locations indicate our localization results.}
		\label{sample_result_plot_map}
	\end{figure*}

	%\smallskip
	Some qualitative results for Pittsburgh, PA are presented in Fig. \ref{sample_result_plot_map}.
	
	%***Sample qualitative result
	%\section{Results on the First Dataset}
	\subsection{Analysis }
	
	\subsubsection{Outlier Handling} In order to show that our dominant set-based feature matching technique is robust in handling outliers, we  conduct  an experiment by fixing the number of NNs (disabling the dynamic selection of NNs) to different numbers. It is obvious that the higher the number of NNs are considered for each query feature, the higher will be the number of outlier NNs in the input graph, besides the increased computational cost and an elevated chance of query features whose NNs do not contain any inliers surviving the pruning stage.

\begin{figure}
		\centering
		\includegraphics[width=1\linewidth, trim=1cm 7.6cm 1cm 8cm, clip ]{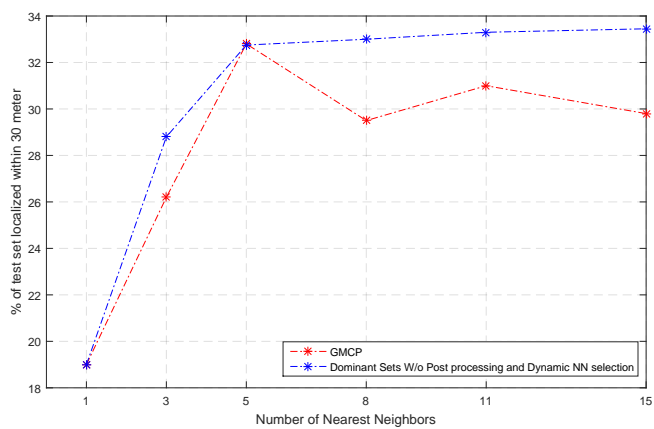}
		\caption{Geo-localization results using different number of NN}
		\label{plot:outlier_handling}
	\end{figure}

	Fig. \ref{plot:outlier_handling} shows the results of geo-localization obtained by using GMCP and dominant set based feature matching on 102K Google street view images \cite{amirshahpami2014}. The graph shows the percentage of the test set localized within the distance of 30 meters as a function of number of NNs. The blue curve shows the results using dominant sets: it is evident that when the number of NNs increases, the performance improves despite the fact that more outliers are introduced in the input graph. This is mainly because our framework takes advantage of the few inliers that are added along with many outliers. The red curve shows the results of GMCP based localization and as the number of NNs increase the results begin to drop. This is mainly due to the fact that their approach imposes hard constraint that at least one matching reference feature should be selected for each query feature whether or not the matching feature is correct.
	
	%\smallskip
	\subsubsection{Effectiveness of the Proposed Post Processing} 
\begin{figure}
		\centering
		\includegraphics[width=1\linewidth, trim=2.00cm 7.1cm 1.75cm 7.4cm, clip ]{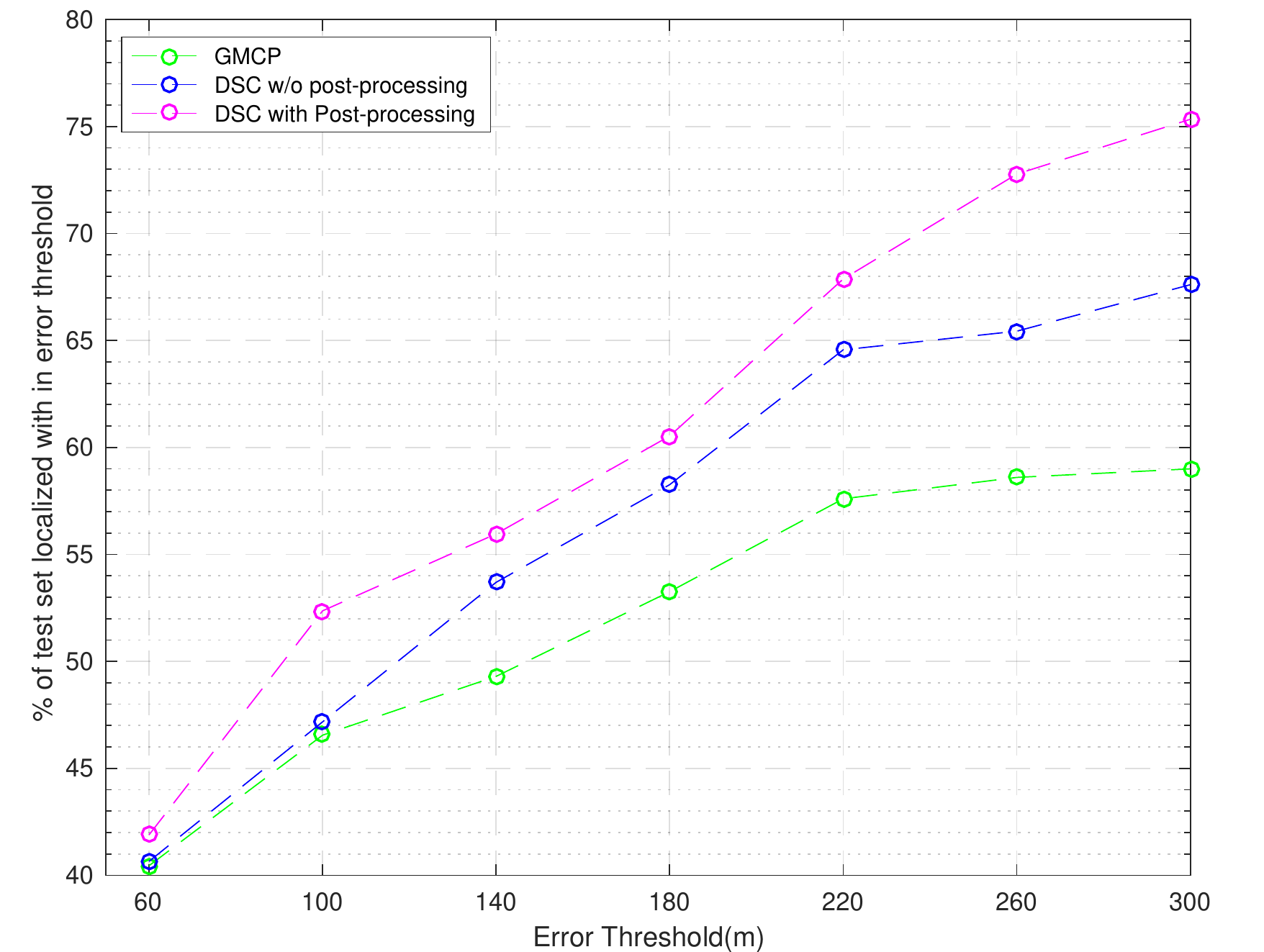}
		\caption{The effectiveness of constrained dominant set based post processing step over simple voting scheme. }
		\label{plot:post_processing}
	\end{figure}

In order to show the effectiveness of the post processing step, we perform an experiment comparing our constrained dominant set based post processing with a simple voting scheme to select the best matching reference image. The GPS information of the best matching reference image is used to estimate the location of the query image. In Fig. \ref{plot:post_processing}, the vertical axis shows the percentage of the test set localized within a particular error threshold shown in the horizontal axis (in meters). The blue and magenta curves depict geo-localization results of our approach using a simple voting scheme and constrained dominant sets, respectively. The green curve shows the results from GMCP based geo-localization. As it is clear from the results, our approach with post processing exhibits superior performance compared to both GMCP and our baseline approach.
	
	Since our post-processing algorithm can be easily plugged in to an existing retrieval methods, we perform another experiment to determine how much improvement we can achieve by our post processing. We use \cite{TorSivOkuPajPAMI2015,ToliasICLR2016,ToliasECCV2016} methods to obtain candidate reference images and employ as an edge weight the similarity score generated by the corresponding approaches. Table \ref{table:PostTable} reports, for each dataset, the first row shows rank-1 result obtained from the existing algorithms while the second row (w\_post) shows rank-1 result obtained after adding the proposed post-processing step on top of the retrieved images. For each query, we use the first 20 retrieved reference images. As the results demonstrate, Table \ref{table:PostTable}, we are able to make up to 7\%  and 4\% improvement on  102k Google street view images and \textit{\textbf{WorldCities}} datasets, respectively. We ought to note that, the total additional time required to perform the above post processing, for each approach, is less than 0.003 seconds on average.
	\begin{figure}
		\centering
		\includegraphics[width=1.02\linewidth, trim=2.00cm 7cm 1.75cm 7.5cm, clip ]{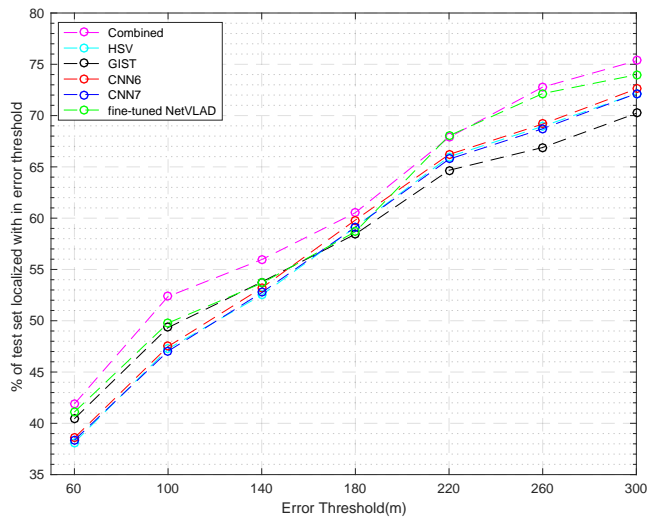}
		\caption{Comparison of geo-localization results using different global features for our post processing step.}
		\label{plot:global_features}
	\end{figure}
	
	%\begin{table}[h]
	%	\centering		
	%	\begin{tabular}{cc|c|c|c|c|r}
	%		\centering
	%		%\cline{3-6}
	%		&	{Methods} &{NetVLAD} & {NetVLAD*} &{RMAC}&{MAC} \\ \cline{1-6}
	%		&	{Rank1} & 49.2 & 56.00 &35.16 & 29.00  \\ \cline{1-6}
	%		&	{Rank1 w\_post} & \textbf{51.60} & \textbf{58.05} &\textbf{40.18} & \textbf{36.30}  \\ \cline{1-6}
	%		
	%		
	%	\end{tabular}
	%	
	%	\caption{Results of the experiment, done on the \textbf{first dataset}, to see the impact of the post-processing step when the candidates of reference images are obtained by other image retrieval algorithms.   %\cite{ToliasICLR2016,ToliasECCV2016}.
	%	}
	%	\label{table:Post_Process1}
	%\end{table}
	
	%\begin{table}[h]
	%	\centering		
	%	\begin{tabular}{cc|c|c|c|c|r}
	%		\centering
	%		%\cline{3-6}
	%		&	{Methods} &{NetVLAD} & {NetVLAD*} &{RMAC}&{MAC} \\ \cline{1-6}
	%		&	{Rank1} & 50.00 & 50.61 &29.96 & 22.87  \\ \cline{1-6}
	%		&	{Rank1 w\_post} & \textbf{53.04} & \textbf{52.23} &\textbf{33.16} & \textbf{26.31}  \\ \cline{1-6}
	%		
	%		
	%	\end{tabular}
	%	
	%	\caption{Results of the experiment, done on the \textbf{second dataset}, to see the impact of the post-processing step when the candidates of reference images are obtained by other image retrieval algorithms %\cite{ToliasICLR2016,ToliasECCV2016}.
	%	}
	%	\label{table:Post_Process2}
	%\end{table}
	
	\begin{table}[h]
		\centering
		\begin{tabular}{cc|c|c|c|c|c|r}
			
			%\cline{3-6}
			&	 &{NetVLAD} & {NetVLAD*} &{RMAC}&{MAC} \\ \cline{1-6}
			%& {Datasets }  &{IDF1}$\uparrow$ & IDP$\uparrow$ &  IDR$\uparrow$ &h\\ \cline{1-6}
			%	& & 2 & 3 & 5 & 7 
			\cline{1-6}
			\multicolumn{1}{ c  }{\multirow{2}{*}{Dts 1} } &
			\multicolumn{1}{ |c| }{Rank1} & 49.2 & 56.00 &35.16 & 29.00  \\ \cline{2-6}
			\multicolumn{1}{ c  }{}                        &
			\multicolumn{1}{ |c| }{w\_post} & \textbf{51.60} & \textbf{58.05} &\textbf{40.18} & \textbf{36.30}  \\ \cline{1-6}
			\cline{1-6}
			\multicolumn{1}{ c  }{\multirow{2}{*}{Dts 2} } &
			%\multicolumn{1}{ |c| }{\cite{RisSolZouCucTomECCV16}} & 56.2 &	67.0 &	48.4&7 \\ \cline{2-6}
			\multicolumn{1}{ |c| }	{Rank1} & 50.00 & 50.61 &29.96 & 22.87  \\ \cline{2-6}
			\multicolumn{1}{ c  }{}                        &
			\multicolumn{1}{ |c| }{ w\_post} & \textbf{53.04} & \textbf{52.23} &\textbf{33.16} & \textbf{26.31}  \\ \cline{1-6}

		\end{tabular}
		
		\caption{Results of the experiment, done on the 102k Google street view images (Dts1) and \textit{\textbf{WorldCities}} (Dts2) datasets, to see the impact of the post-processing step when the candidates of reference images are obtained by other image retrieval algorithms }
		\label{table:PostTable}
	\end{table}

	The NetVLAD results are obtained from the features generated using the best trained model downloaded from the author’s project page \cite{TorSivOkuPajPAMI2015}. It's fine-tuned version (NetVLAD*) is obtained from the model we fine-tuned using images within 24m range as a positive set and images with GPS locations greater than 300m  as a negative set.
	
	The MAC and RMAC results are obtained using MAC and RMAC representations extracted from fine-tuned VGG networks downloaded from the authors webpage \cite{ToliasICLR2016,ToliasECCV2016}.
	
	%to choose the final best matching reference image. The green curve shows the localization results after post processing step. The superior performance of our post processing step is mainly due to  ..............
	%%trim=2cm(left) 3cm(bottom) 1.75cm(right) 2.5cm(top),clip
	
	%\vspace*{-1.25cm}
	\smallskip
	\subsubsection{Assessment of Global Features Used in Post Processing Step} The input graph for our post processing step utilizes the global similarity between the query and the matched reference images. Wide variety of global features can be used for the proposed technique. In our experiments, the similarity between query and the corresponding matched reference images is computed between their global features, using HSV, GIST, CNN6, CNN7 and fine-tuned NetVLAD. The performance of the proposed post processing technique highly depends on the discriminative ability of the global features used to built the input graph.
	
	Depending on how informative the feature is, we dynamically assign a weight for each global feature based on the area under the normalized score between the query and the matched reference images. % Note that in this way for each query different weights will be used to combine the features.
	To show the effectiveness of %merging global features rather than taking individual once,
	this approach, we perform an experiment to find the location of our test set images using both individual and combined global features. Fig. \ref{plot:global_features} shows the results attained by using fine-tuned NetVLAD, CNN7, CNN6, GIST, HSV and by combining them together. The combination of all the global features outperforms the individual feature performance, demonstrating the benefits of fusing the global features based on their discriminative abilities for each query.
	
	%\subsubsection{Impact of the post-processing step }

	%***************************
	\section{Conclusion and Future Work}\label{secConclusion}
	In this paper, we proposed a novel framework for city-scale image geo-localization. Specifically, we introduced dominant set clustering-based multiple NN feature matching approach. Both global and local features are used in our matching step in order to improve the matching accuracy. In the experiments, carried out on two large city-scale datasets, we demonstrated the effectiveness of  post processing employing the novel constrained dominant set over a simple voting scheme. Furthermore, we showed that our proposed approach is 200 times, on average, faster than GMCP-based approach \cite{amirshahpami2014}. Finally, the newly-created dataset (\textit{\textbf{WorldCities}}) containing more than 300k Google Street View images used in our experiments will be made available to the public for research purposes.
	
	As a natural future direction of research, we can extend the results of this work for estimating the geo-spatial trajectory of a video in a city-scale urban environment from a moving camera with unknown intrinsic camera parameters.

%\smallskip
%\noindent \textbf{Acknowledgment:}

\bibliographystyle{IEEETran} %IEEETran
%\begin{thebibliography}{1}
\bibliography{GeoLocDomSetV1}

%\end{thebibliography}
%\newpage
\vspace*{-2cm}
\begin{IEEEbiography}[{\includegraphics[width=1.0in,height=1.5in,trim=6cm 3cm 6cm 0.5cm,clip,keepaspectratio]{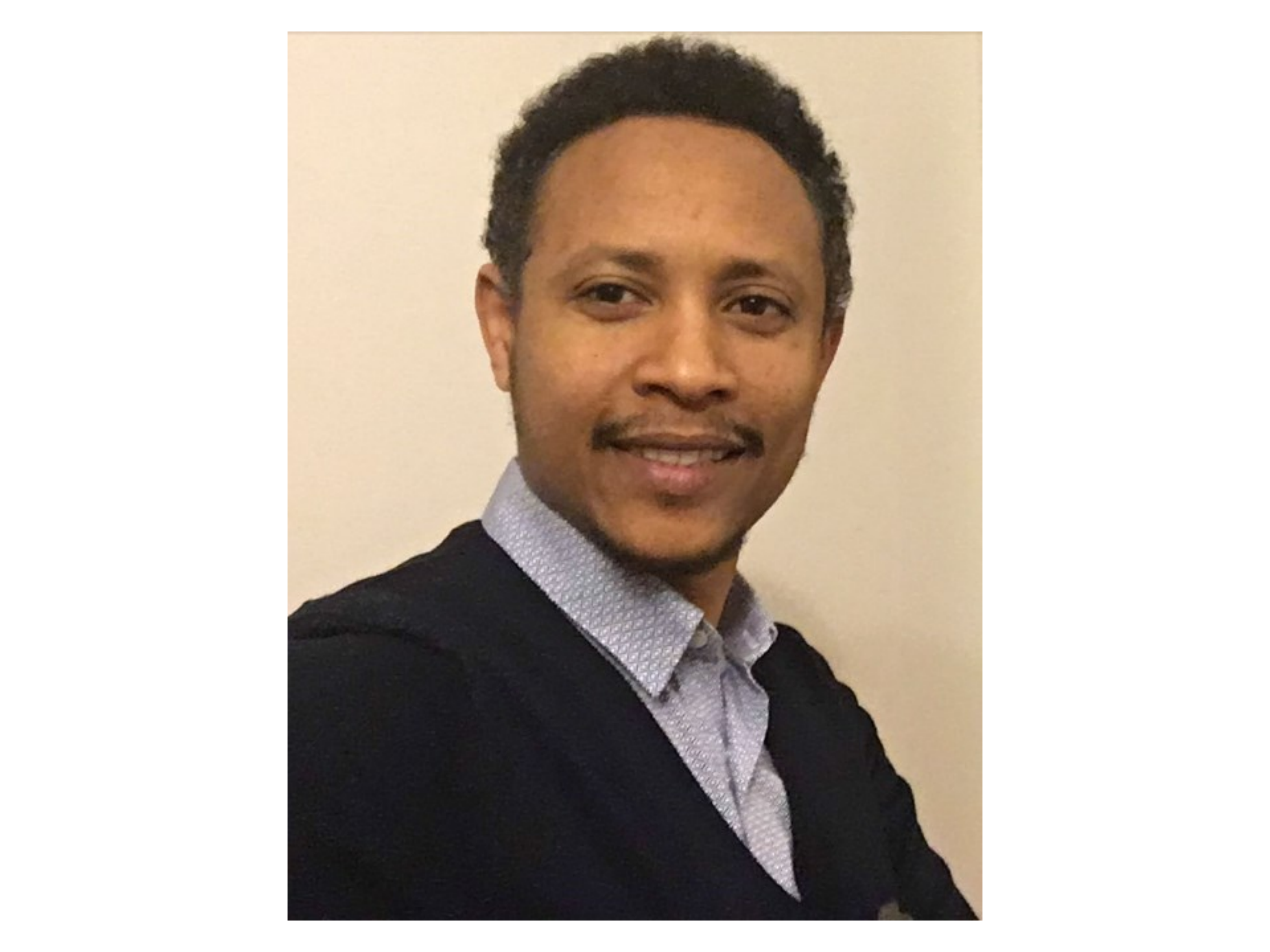}}]{Eyasu Zemene}
received the BSc degree in Electrical Engineering from Jimma University in 2007, he then worked in Ethio Telecom for 4 years till he joined CaFoscari University (October 2011) where he got his MSc in Computer Science in June 2013. September 2013, he won a 1 year research fellow to work on Adversarial Learning at Pattern Recognition and Application lab of University of Cagliari. Since September 2014 he is a PhD student of CaFoscari University under the supervision of prof. Pelillo. Working towards his Ph.D. he is trying to solve different computer vision and pattern recognition problems using theories and mathematical tools inherited from graph theory, optimization theory and game theory. Currently, Eyasu, as part of his PhD, is working as a research assistant at Center for Research in Computer Vision at University of Central Florida under the supervision of Dr. Mubarak Shah. His research interests are in the areas of Computer Vision, Pattern Recognition, Machine Learning, Graph theory and Game theory.
\end{IEEEbiography}\vspace*{-1.2cm}
\begin{IEEEbiography}[{\includegraphics[width=1.0in,height=1.5in,clip,keepaspectratio]{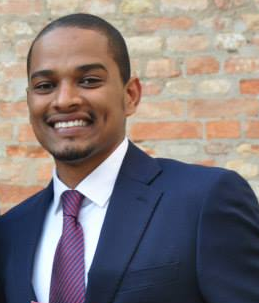}}]{Yonatan Tariku}
received his BSc degree in computer science from Arba Minch University in 2007. He has worked 5 years at Ethio-telecom as senior programmer and later joined CaFoscari University of Venice and received his MSc degree (with Honor) in computer science in 2014. He is currently a PhD student at IUAV university of Venice starting from 2014. He is now a research assistance, towards his PhD, at Center for Research in Computer Vision at University of Central Florida. His research interests include multi-target tracking, segmentation, image and video geo-localization, game theoretic model and graph theory.
\end{IEEEbiography}\vspace*{-2cm}
\begin{IEEEbiography}[{\includegraphics[width=1.0in,height=1.5in,clip,keepaspectratio]{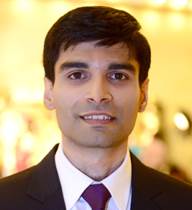}}]{Haroon Idrees}
is a Postdoctoral Associate at the Center for Research in Computer Vision at University of Central Florida. He received the BSc (Hons) degree in Computer Engineering from the Lahore University of Management Sciences, Pakistan in 2007, and the PhD degree in Computer Science from the University of Central Florida in 2014. He has published several papers in conferences and journals such as CVPR, ECCV, Journal of Image and Vision Computing, and IEEE Transactions on Pattern Analysis and Machine Intelligence. His research interests include crowd analysis, object detection and tracking, wide area analysis, multi-camera and airborne surveillance, and multimedia content analysis.
\end{IEEEbiography}\vspace*{-2cm}
\begin{IEEEbiography}[{\includegraphics[width=1.0in,height=1.5in,clip,keepaspectratio]{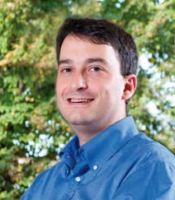}}]{Andrea Prati} Andrea Prati graduated in Computer Engineering at the University of Modena and Reggio Emilia in 1998. He got his PhD in Information Engineering in 2002 from the same University. After some post-doc position at University of Modena and Reggio Emilia, he was appointed as Assistant Professor at the Faculty of Engineering of Reggio Emilia (University of Modena and Reggio Emilia) from 2005 to 2011, and then as Associate Professor at the Department of Design and Planning in Complex Environments of the University IUAV of Venice, Italy. In 2013 he has been promoted to full professorship, waiting for official hiring in the new position. In December 2015 he moved to the Department of Engineering and Architecture of the University of Parma. Author of 7 book chapters, 31 papers in international referred journals (including 9 papers published in IEEE Transactions) and more than 100 papers in proceedings of international conferences and workshops. Andrea Prati is Senior Member of IEEE, Fellow of IAPR ("For contributions to low- and high-level algorithms for video surveillance"), and member of GIRPR.
\end{IEEEbiography}\vspace*{-1.9cm}
\begin{IEEEbiography}[{\includegraphics[width=1.0in,height=1.5in,clip,keepaspectratio]{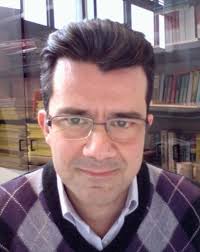}}]{Marcello Pelillo} is Full Professor of Computer Science at CaFoscari University in Venice, Italy, where he directs the European Centre for Living Technology (ECLT) and leads the Computer Vision and Pattern Recognition group. He held visiting research positions at Yale University, McGill University,the University of Vienna, York University (UK), the University College London, and the National ICT Australia (NICTA). He has published more than 200 technical papers in refereed journals, handbooks, and conference proceedings in the areas of pattern recognition, computer vision and machine learning. He is General Chair for ICCV 2017 and has served as Program Chair for several conferences and workshops (EMMCVPR, SIMBAD, S+SSPR, etc.). He serves (has served) on the Editorial Boards of the journals IEEE Transactions on Pattern Analysis and Machine Intelligence (PAMI), Pattern Recognition, IET Computer Vision, Frontiers in Computer Image Analysis, Brain Informatics, and serves on the Advisory Board of the International Journal of Machine Learning and Cybernetics. Prof. Pelillo is a Fellow of the IEEE and a Fellow of the IAPR.
\end{IEEEbiography} \vspace*{-1.6cm}
\begin{IEEEbiography}[{\includegraphics[width=1.0in,height=3in,trim=0cm 0cm 0cm 0cm, clip,keepaspectratio]{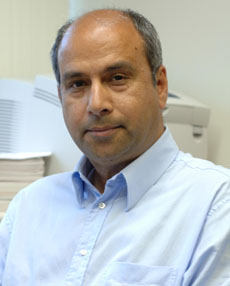}}]{Mubarak Shah,} the Trustee chair professor of computer science, is the founding director of the Center for Research in Computer Vision at the University of Central Florida (UCF). He is an editor of an international book series on video computing, editor-in-chief of Machine Vision and Applications journal, and an associate editor of ACM Computing Surveys journal. He was the program cochair of the IEEE Conference on Computer Vision and Pattern Recognition (CVPR) in 2008, an associate editor of the IEEE Transactions on Pattern Analysis and Machine Intelligence, and a guest editor of the special issue of the International Journal of Computer Vision on Video Computing. His research interests include video surveillance, visual tracking, human activity recognition, visual analysis of crowded scenes, video registration, UAV video analysis, and so on. He is an ACM distinguished speaker. He was an IEEE distinguished visitor speaker for 1997-2000 and received the IEEE Outstanding Engineering Educator Award in 1997. In 2006, he was awarded a Pegasus Professor Award, the highest award at UCF. He received the Harris Corporations Engineering Achievement Award in 1999, TOKTEN awards from UNDP in 1995, 1997, and 2000, Teaching Incentive Program Award in 1995 and 2003, Research Incentive Award in 2003 and 2009, Millionaires Club Awards in 2005 and 2006, University Distinguished Researcher Award in 2007, Honorable mention for the ICCV 2005 Where Am I? Challenge Problem, and was nominated for the Best Paper Award at the ACM Multimedia Conference in 2005. He is a fellow of the IEEE, AAAS, IAPR, and SPIE.
\end{IEEEbiography}
%\enlargethispage{-2cm}\vfill
\end{document}